\documentclass[lettersize,journal]{IEEEtran}
\usepackage{amsmath,amsfonts}
\usepackage{algorithmic}
\usepackage{algorithm}
\usepackage{array}
\usepackage[caption=false,font=normalsize,labelfont=sf,textfont=sf]{subfig}
\usepackage{textcomp}
\usepackage{stfloats}
\usepackage{url}
\usepackage{verbatim}
\usepackage{graphicx}
\usepackage{cite}
\usepackage{booktabs}
\usepackage{float}
\usepackage{xcolor}
\usepackage{ragged2e}   % 用于文本对齐控制
\begin{document}

\title{Development of a 15-Degree-of-Freedom Bionic Hand with Cable-Driven Transmission and Distributed Actuation}

\author{Haoqi Han, Yi Yang, Yifei Yu, Yixuan Zhou, Xiaohan Zhu, Hesheng Wang,~\IEEEmembership{Senior Member,~IEEE}
        % <-this % stops a space
\thanks{
        Manuscript received Month xx, 2xxx; revised Month xx, xxxx; accepted Month x, xxxx.
        This work was supported in part by the Natural Science Foundation of China under Grant 62225309, U24A20278, 62361166632, U21A20480 and 62403311. (Corresponding Author: Hesheng Wang)

        Haoqi Han is with the Department of Computer Science and Engineering, Shanghai Jiao Tong University, Shanghai, 200240, China (e-mail: hhq123@sjtu.edu.cn).

        Yi Yang is with the Department of Engineering Mechanics, Shanghai Jiao Tong University, Shanghai, 200240, China (e-mail: yangyi04@sjtu.edu.cn).

        Yifei Yu is with the Department of Automation Engineering,Shanghai University of Electric Power,Shanghai,China(e-mail:yuyifei@mail.shiep.edu.cn)

        Yixuan Zhou, Xiaohan Zhu are with the Department of Automation, Shanghai Jiao Tong University, Shanghai, 200240, China (e-mail: yixuanzhou@sjtu.edu.cn, zxh\_2001@sjtu.edu.cn).

        Hesheng Wang is with the Department of Automation, Key Laboratory of System Control and Information Processing of Ministry of Education, Key Laboratory of Marine Intelligent Equipment and System of Ministry of Education, Shanghai Engineering Research Center of Intelligent Control and Management, Shanghai Jiao Tong University, Shanghai, 200240, China (e-mail: wanghesheng@sjtu.edu.cn).
    }
\thanks{Manuscript received March 1, 2025; revised August 16, 2025.}}

\markboth{Journal of \LaTeX\ Class Files,~Vol.~14, No.~8, August~2025}%
{Shell \MakeLowercase{\textit{et al.}}: A Sample Article Using IEEEtran.cls for IEEE Journals}

\IEEEpubid{0000--0000/00\$00.00~\copyright~2025 IEEE}

\maketitle

\begin{abstract}

In robotic hand research, minimizing the number of actuators while maintaining human-hand-consistent dimensions and degrees of freedom constitutes a fundamental challenge. Drawing bio-inspiration from human hand kinematic configurations and muscle distribution strategies, this work proposes a novel 15-DoF dexterous robotic hand, with detailed analysis of its mechanical architecture, electrical system, and control system. The bionic hand employs a new tendon-driven mechanism, significantly reducing the number of motors required by traditional tendon-driven systems while enhancing motion performance and simplifying the mechanical structure. This design integrates five motors in the forearm to provide strong gripping force, while ten small motors are installed in the palm to support fine manipulation tasks. Additionally, a corresponding joint sensing and motor driving electrical system was developed to ensure efficient control and feedback. The entire system weighs only 1.4kg, combining lightweight and high-performance features. Through experiments, the bionic hand exhibited exceptional dexterity and robust grasping capabilities, demonstrating significant potential for robotic manipulation tasks.
% 在机器人手的研究中，如何在保持与人手尺寸和自由度（DoF）一致的同时，最大限度地减少执行器的数量，是一项根本性的挑战。本研究从人手运动学配置和肌肉分布策略中汲取生物灵感，提出了一种新型的15自由度灵巧机器人手，并对其机械结构、电气系统和控制系统进行了详细分析。仿生手采用了一种新型肌腱驱动机制，在显著减少传统肌腱驱动系统所需电机数量的同时，提升了运动性能并简化了机械结构。该设计在前臂中集成了五个电机，以提供强大的抓握力，同时在手掌中安装了十个小型电机，以支持精细操作任务。此外，还开发了相应的关节传感和电机驱动电气系统，以确保高效的控制和反馈。整个系统重量仅为1.4千克，兼具轻便和高性能的特点。通过实验，仿生手表现出卓越的灵巧性和强大的抓握能力，在机器人操作任务中展现出巨大潜力。

\end{abstract}


\section{Introduction}
\IEEEPARstart{T}{he}  development of actuator systems with human-level dexterity presents significant challenges \cite{piazza2019century, nguyen2023bioinspiration}, stemming from the bio-integrated nature of the human hand: it is not an isolated entity but a highly coupled system intricately connected through skeletal-muscular-neural networks to the forearm, forming a synergistic functional unit. This anatomical architecture enables both high-power grasping and fine manipulation with sub-millimeter precision. Furthermore, the human hand exhibits multimodal sensing capabilities for force, temperature, and texture discrimination, demonstrating exceptional environmental adaptability and task generalization. This demands biomimetic actuators to simultaneously achieve kinematic dexterity, power density, and proprioceptive integration within constrained volumes, creating a characteristic trilemma in mechatronic system design.
% {T}具有人类水平灵巧性的执行器系统的开发面临着重大挑战\cite{piazza2019century, nguyen2023bioinspiration}，这源于人手的生物集成特性：它不是一个孤立的实体，而是一个高度耦合的系统，通过骨骼-肌肉-神经网络与前臂错综复杂地相连，形成一个协同工作的功能单元。这种解剖结构既能够实现高功率的抓取，也能以亚毫米的精度进行精细操作。此外，人手还具有多模态感知能力，能够感知力、温度和纹理差异，表现出卓越的环境适应性和任务泛化能力。这就要求仿生执行器在有限的体积内同时实现运动灵巧性、功率密度和本体感觉集成，这在机电一体化系统设计中构成了一个典型的三难问题

\begin{figure}
    \centering
    \includegraphics[width=0.8\linewidth]{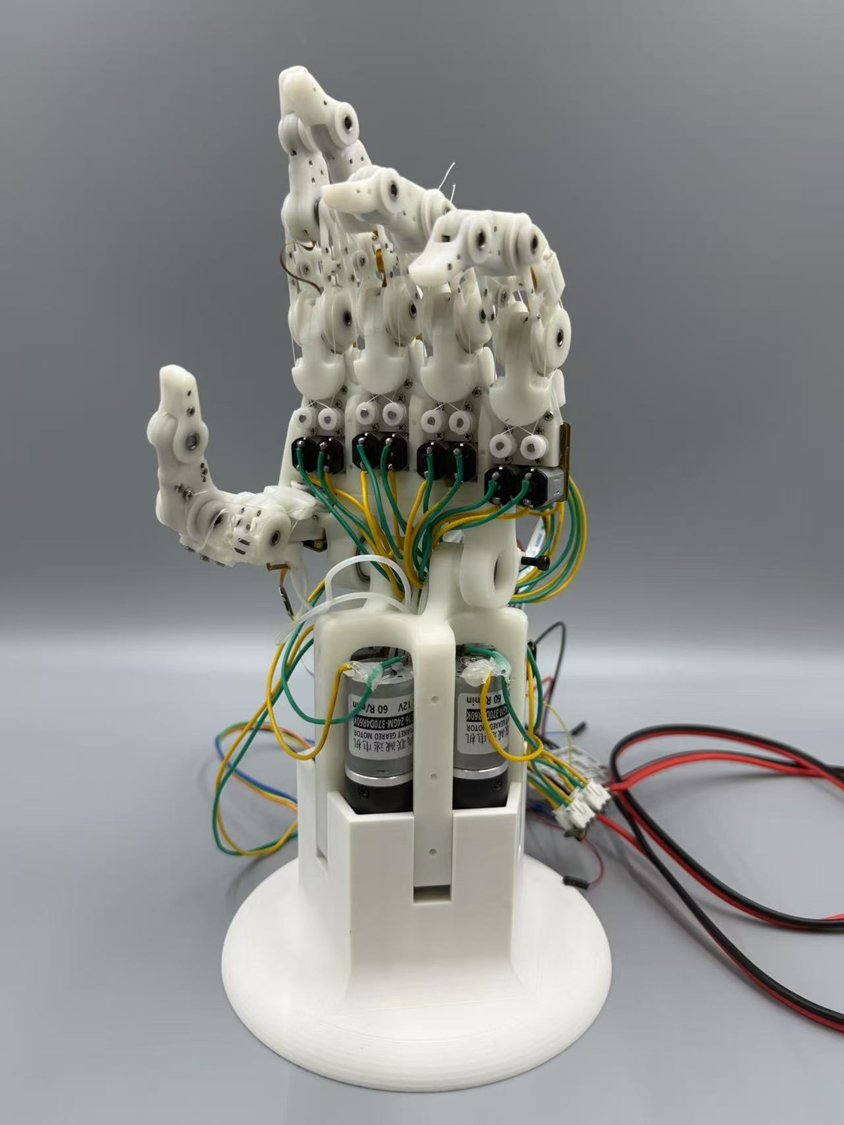}
    \caption{The IRMV Hand}
    \label{fig:physical}
\end{figure}

The motion of the human finger is primarily defined by the MCP (metacarpophalangeal), PIP (proximal interphalangeal), and DIP (distal interphalangeal) joints. The movements of the PIP and DIP joints are relatively straightforward, typically driven by the coordinated action of the extensor muscles, deep flexor muscles, and superficial flexor muscles, with these movements often being coupled. In contrast, the MCP joint involves more complex motions, such as abduction and adduction, which are controlled by the interosseous muscles, lumbrical muscles, and both the deep and superficial flexors.\cite{vonschroeder2001anatomy}
% 人类手指的运动主要由掌指关节（MCP）、近端指间关节（PIP）和远端指间关节（DIP）来定义。近端指间关节和远端指间关节的运动相对直接，通常由伸肌、深部屈肌和浅部屈肌的协同作用驱动，且这些运动往往是相互耦合的。相比之下，掌指关节的运动更为复杂，如外展和内收，这些运动由骨间肌、蚓状肌以及深部和浅部屈肌共同控制。[vonschroeder2001anatomy]

Biomimetic finger width design presents a fundamental challenge. Universal human devices strictly adhere to hand anatomical dimensions: the proximal phalanx width measures 16.4±1.4 mm for adult males and 14.8±1.2 mm for females. This dimensional constraint directly governs interaction object design:
% 仿生手指宽度设计面临着一项根本性挑战。通用人体设备严格遵循手部解剖尺寸：成年男性的近节指骨宽度为16.4±1.4毫米，女性为14.8±1.2毫米。这一尺寸限制直接决定了交互对象的设计：
\begin{enumerate}
    \item \textbf{Precision tools:} Pen barrel diameter (7–11  mm), USB-A connector (12 mm), occupying 40\%–60\% of human finger width to optimize the compromise between tactile perception and frictional control
    \item \textbf{Grasping objects:} Tool handle diameter (20–35 mm), dumbbell grip (28–32 mm), constituting 10\%–20\% of proximal phalanx length (72.0 ± 8.0 mm in males), achieving co-optimization of ergonomic comfort and payload capacity under minimal moment arms
\end{enumerate}

% {精密工具：}笔杆直径（7-11毫米），USB-A接口（12毫米），占据人体手指宽度的40%-60%，以优化触觉感知与摩擦控制之间的平衡
% {抓取物体：}工具手柄直径（20-35毫米），哑铃握把（28-32毫米），占近节指骨长度（男性为72.0±8.0毫米）的10%-20%，在最小力臂下实现人体工程学舒适度和有效载荷能力的共同优化

In addition, dexterity is a central focus in the development of bionic hands for robotic manipulation tasks. The degrees of freedom in a bionic hand are crucial for enabling complex operations. For instance, to rotate lightweight objects at arbitrary angles within the palm, a bionic hand requires at least four sets of two-link systems, each necessitating a minimum of 3 DOF. Moreover, for grasping tasks, the flexion of the bionic hand’s fingers must be capable of delivering substantial power output. This requires at least 12 DOF to meet basic operational requirements.
% 此外，在开发用于机器人操作任务的仿生手时，灵巧性是一个核心关注点。仿生手的自由度对于实现复杂操作至关重要。例如，要在手掌内以任意角度旋转轻质物体，仿生手至少需要四组两连杆系统，每组至少需要3个自由度（DOF）。此外，对于抓取任务，仿生手的手指屈曲必须能够提供相当大的输出力。这至少需要12个自由度才能满足基本操作要求。

The rapid evolution of dexterous manipulation research in recent years has demonstrated the potential of high-DoF bionic hands in human-like operational tasks \cite{suresh2023neural}. However, predominant designs with average finger widths \(\approx \)30 mm constrain demonstration objects to model props. While such validations confirm algorithmic feasibility, they fail to achieve physical interaction with human-centric devices. Consequently, compressing key dimensions of bionic hands to anatomically matched ranges while preserving equivalent kinematics, thereby enabling universal operation of human tools, constitutes a critical technical challenge in the field.
% 近年来，灵巧操作研究的迅速发展展示了高自由度仿生手在类人操作任务中的潜力\cite{suresh2023neural}。然而，主流设计的手指宽度平均约为30毫米，这限制了演示对象只能使用模型道具。虽然此类验证证实了算法的可行性，但它们无法实现与以人为本的设备的物理交互。因此，在保持等效运动学特性的同时，将仿生手的关键尺寸压缩到与解剖学相匹配的范围，从而实现人类工具的通用操作，是该领域面临的一项关键技术挑战。

In the design of bionic hands, cable-driven mechanisms offer more flexible configurations. There are generally two types of cable-driven bionic hands: tendon-driven and independently actuated. The tendon-driven bionic hand uses multiple motors to collaboratively drive the finger joints, employing cables to mimic the way muscles drive the skeleton. However, this approach increases the size of the device. In contrast, the independently actuated bionic hand uses a separate motor for each joint, matching the number of joints, which simplifies the transmission structure.
% 在仿生手的设计中，绳索驱动机构提供了更为灵活的配置。绳索驱动仿生手通常有两种类型：肌腱驱动型和独立驱动型。肌腱驱动型仿生手利用多个电机协同驱动手指关节，通过绳索来模拟肌肉驱动骨骼的方式。然而，这种方法增加了设备的体积。相比之下，独立驱动型仿生手为每个关节配备一个单独的电机，与关节数量相匹配，从而简化了传动结构。

For actuators, the efficiency is typically conserved. When a bionic hand is required to perform both grasping and manipulation tasks, larger actuators are necessary, which inevitably increases the size. However, bionic hands are generally expected to be lightweight and agile, and overly heavy drive systems contradict this goal. To address this, we adopt a distributed drive system design: a cluster of smaller motors is placed in the palm to assist with manipulation, while the core force needed for grasping is provided by a forearm motor. This design balances the demands of grasping and manipulation tasks while preserving the lightness and agility of the bionic hand.
% 对于执行器而言，效率通常是保守的。当仿生手需要执行抓取和操作任务时，需要更大的执行器，这不可避免地增加了尺寸。然而，仿生手通常被期望为轻便且灵活，而过重的驱动系统与这一目标相悖。为了解决这一问题，我们采用了分布式驱动系统设计：一组较小的电机放置在手掌中以辅助操作，而抓取所需的核心力量则由前臂电机提供。这种设计在满足抓取和操作任务需求的同时，保持了仿生手的轻便和灵活性。

This paper proposes a design for a fully actuated bionic hand, as shown in Fig. \ref{fig:physical}. The main feature of this design is its distributed drive system, with 10 motors for manipulation mounted in the palm and an additional 5 motors located in the forearm. Additionally, we introduce a novel single-finger, three-degree-of-freedom parallel transmission mechanism, which simulates human hand movement with advantages such as simplicity and a wide working range. To ensure performance, corresponding sensors and motor driving circuits are designed. The main features can be summarized as follows:
% 本文提出了一种全驱动仿生手的设计，如图\ref{fig:physical}所示。该设计的主要特点是其分布式驱动系统，其中10个用于操纵的电机安装在手掌中，另外5个电机位于前臂。此外，我们还引入了一种新颖的单指三自由度并联传动机构，该机构模拟人手运动，具有结构简单和工作范围广等优点。为确保性能，设计了相应的传感器和电机驱动电路。其主要特点可概括如下：

\begin{enumerate}
    \item \textbf{Distributed Drive Design}: Motors are distributed across the forearm and palm, optimizing spatial utilization and creating a more compact structure.
    \item \textbf{3-DoF Tendon Transmission Mechanism}: A highly efficient parallel tendon transmission design enhances both transmission efficiency and motion flexibility.
    \item \textbf{Bionic Hand Hardware System}: A new hardware system has been developed to achieve efficient motion control and precise operation of the bionic hand.
\end{enumerate}
% {分布式驱动设计}：电机分布于前臂和手掌各处，优化了空间利用率，使结构更加紧凑。
% {3自由度肌腱传动机构}：高效并联肌腱传动设计，既提高了传动效率，又增强了运动灵活性。
% {仿生手硬件系统}：为实现对仿生手的高效运动控制和精确操作，我们开发了一种新的硬件系统。

The rest of this paper is organized as follows: Section II introduces mainstream technical approaches for anthropomorphic bionic hands. Section III discusses the mechanical transmission system of the bionic hand, the distribution of power sources, and the design methods for control circuits. Section IV analyzes the kinematics, dynamics, and working space of the bionic hand, and presents its control principles. Section V presents experimental results demonstrating the bionic hand’s performance in grasping tasks. Finally, the paper concludes with a summary of the proposed design and future research directions.

% 本文其余部分结构如下：第二节讨论仿生手的机械传动系统、动力源的分布以及控制电路的设计方法。第三节分析仿生手的动力学、动力学和工作空间，并介绍其控制原理。第四节展示实验结果，演示仿生手在抓取任务中的表现。最后，本文总结了所提出的设计和未来的研究方向。

\section{Related Works}

For a long time, humans have sought to replicate the physiological structure of the human hand, aiming to develop bionic hands for use in fields such as service, industry, and rescue operations. These bionic hands provide efficient alternatives to relieve humans from repetitive labor or hazardous environments. However, designing a fully dexterous bionic hand presents a significant engineering challenge. The human hand’s functionalities, including individual finger movement, grasping, oppositional configuration changes, and tactile perception, require integration of various transmission mechanisms, actuators, and motion sensors. Additionally, challenges related to spatial arrangement, energy efficiency, and control systems must be addressed.
% 长期以来，人类一直试图复制人手生理结构，旨在开发出仿生手，用于服务、工业和救援等领域。这些仿生手为人类提供了高效的替代方案，使人们能够从重复性劳动或危险环境中解脱出来。然而，设计一款完全灵巧的仿生手是一项重大的工程挑战。人手的功能，包括单个手指的运动、抓握、对掌构型变化和触觉感知，需要集成各种传动机构、执行器和运动传感器。此外，还必须解决与空间布局、能效和控制系统相关的挑战。

Current approaches to building such bionic hands include module-driven mechanisms, linkage-driven mechanisms, and tendon-driven mechanisms. Each approach has distinct advantages and challenges: direct-drive motor systems offer rapid response but are bulky; linkage-driven mechanisms provide stability but are complex; tendon-driven systems are flexible and lightweight but require high-quality materials and precise control.
% 目前构建仿生手的方法包括模块驱动机构、连杆驱动机构和肌腱驱动机构。每种方法都有其独特的优势和挑战：直接驱动电机系统响应迅速但体积庞大；连杆驱动机构稳定性好但结构复杂；肌腱驱动系统灵活轻便但需要高质量材料和精确控制。

\subsubsection{Module-Driven} The module-driven mechanisms is one of the most effective methods for constructing bionic hands, where servo motors are directly integrated within the fingers. This approach offers several benefits, including simplified design, ease of maintenance, and stable control. Representative designs include the KITECH Hand \cite{bae2012development, lee2017kitechhand} , Allegro Hand \cite{allegrohand}, Leap Hand \cite{shaw2023leap}. However, direct integration of servo motors in the fingers increases motion inertia, compromising dynamic response and flexibility. Additionally, limited space within the fingers restricts the power output, hindering performance in grasping and manipulation tasks.  On the other hand, module-based biomimetic hands typically exhibit substantial dimensions, with an average finger width of approximately 30mm. This necessitates the use of larger-than-conventional objects during operational demonstrations—objects that exceed the size range typically manipulated by human hands. Thus, this method faces challenges in achieving high performance and compact design. 
% {模块驱动} 模块驱动机制是构建仿生手最有效的方法之一，其中伺服电机直接集成在手指内。这种方法具有多项优势，包括设计简化、维护简便以及控制稳定。 代表性的设计包括KITECH手部\cite{bae2012development, lee2017kitechhand}、Allegro手部\cite{allegrohand}和Leap手部\cite{shaw2023leap}。然而，在手指中直接集成伺服电机会增加运动惯性，从而影响动态响应和灵活性。此外，手指内部空间有限，限制了功率输出，阻碍了抓取和操纵任务的性能。
% 另一方面问题，模块化仿生手通常尺寸较为庞大，其指宽平均设计在30mm附近，使得其展示操作物体通常大于常规人类使用物品，因此，这种方法在实现高性能和紧凑设计方面面临挑战。

\subsubsection{Linkage-Driven} To enhance the compactness of the bionic finger, the power source is often positioned in the palm or forearm, reducing the finger’s size and weight. However, this design increases transmission system complexity, often requiring gears, linkages, or tendon-like mechanisms for power transmission\cite{kawasaki2002dexterous, hu2024variable, li2024multisensory}. For example, the ILDA bionic hand\cite{kim2021integrated} uses a linkage or gear-driven mechanism, offering stability and reducing finger motion inertia. Despite these advantages, linkage or gear mechanisms are not well-suited for long-distance or complex transmission tasks. Consequently, while finger size and weight are optimized, the motor system must still be integrated into the palm, where limited space restricts layout and power output, complicating the balance between force and speed, thus limiting high-performance applications.
% {连杆驱动} 为了提高仿生手指的紧凑性，动力源通常置于手掌或前臂中，从而减小手指的尺寸和重量。然而，这种设计增加了传动系统的复杂性，通常需要齿轮、连杆或肌腱状机构来进行动力传输\cite{kawasaki2002dexterous, hu2024variable, li2024multisensory}。例如，ILDA仿生手\cite{kim2021integrated}采用连杆或齿轮驱动机构，既提供了稳定性，又降低了手指运动的惯性。尽管具有这些优点，但连杆或齿轮机构并不适合长距离或复杂的传动任务。因此，在优化手指尺寸和重量的同时，电机系统仍必须集成到手掌中，而手掌中有限的空间限制了布局和功率输出，使得力与速度之间的平衡变得复杂，从而限制了高性能应用。

\subsubsection{Tendon-Driven} The tendon-driven bionic hand can replicate human hand motion patterns, providing enhanced manipulation and grasping capabilities\cite{deshpande2013mechanisms}. By using synthetic fiber rope, the tendon-driven system can perform complex motion synthesis. Its transmission flexibility allows the power source to be mounted on the robot’s forearm, reducing the size and weight of the fingers. While this design results in a larger overall volume compared to direct-drive or linkage/gear systems, it offers significant advantages in power output and motion flexibility at the finger level. The tendon-driven system enables higher force output and more precise task execution, making it an effective solution for high-performance bionic hand development. Examples include the Shadow Hand, which features 25 movable joints and 20 degrees of freedom, powered by 20 integrated motors and a cable-driven system\cite{shadowhand}. The DLR Hand mimics human hand movement with 19 degrees of freedom, using antagonistic motors and synthetic tendons for a highly realistic model\cite{liu2008modular, chalon2010thumb, friedl2011fas, grebenstein2011dlr, friedl2015frcef}. The Robonaut 2 Hand, developed by NASA, combines DC motors and synthetic tendons to replicate human anatomy, allowing for tool manipulation in space\cite{bridgwater2012robonaut}. The UB Hand from the University of Bologna employs a spring-based structure to replicate finger flexion, offering a simplified but efficient design\cite{lotti2005development, melchiorri2013development}. Lastly, the FLLEX Hand uses 3D printing and flexible materials to create a robust three-degree-of-freedom hand, capable of high fingertip forces, though its performance in finger extension is limited by the spring mechanism.\cite{kim2019fluid} These developments demonstrate the growing potential of bionic hands in high-precision fields.
% {肌腱驱动} 肌腱驱动仿生手能够复制人类手部的运动模式，提供增强的操控和抓握能力\cite{deshpande2013mechanisms}。通过使用合成纤维绳索，肌腱驱动系统能够执行复杂的运动合成。其传动灵活性使得动力源可以安装在机器人的前臂上，从而减小手指的尺寸和重量。虽然这种设计相较于直接驱动或连杆/齿轮系统导致整体体积更大，但在手指层面的动力输出和运动灵活性方面具有显著优势。肌腱驱动系统能够实现更高的力输出和更精确的任务执行，是高性能仿生手开发的有效解决方案。例如，Shadow Hand仿生手具有25个可移动关节和20个自由度，由20个集成电机和一个线缆驱动系统提供动力\cite{shadowhand}。DLR Hand仿生手模仿人类手部运动，具有19个自由度，使用对置电机和合成肌腱打造出高度逼真的模型\cite{liu2008modular, chalon2010thumb, friedl2011fas, grebenstein2011dlr, friedl2015frcef}。NASA开发的Robonaut 2 Hand仿生手结合直流电机和合成肌腱来复制人体解剖结构，从而实现在太空中的工具操控\cite{bridgwater2012robonaut}。博洛尼亚大学的UB Hand仿生手采用基于弹簧的结构来复制手指屈曲动作，提供了一种简化但高效的设计\cite{lotti2005development, melchiorri2013development}。最后，FLLEX Hand仿生手采用3D打印和柔性材料打造出一个坚固的三自由度手部，能够产生高指尖力，尽管其手指伸展性能受到弹簧机构的限制\cite{kim2019fluid}。这些发展展示了仿生手在高精度领域日益增长的潜力。

However, despite the clean and organized design of integrating motor systems in the forearm, this solution presents challenges in practical applications, particularly for humanoid robots\cite{min2021development, melchiorri2013development}. The large gripping force relies on forearm muscles for power support, while fine manipulation is achieved through precise control in the palm. Therefore, bionic hand design must balance force output with flexibility, integrating forearm power sources and palm precision for efficient, multifunctional performance. This layered drive design not only more closely resembles the physiological structure of the human hand but also provides a feasible solution for practical bionic hand applications\cite{kim2014roboray, weng2025bidexhand}.
% 然而，尽管前臂集成电机系统的设计简洁有序，但这种解决方案在实际应用中仍面临挑战，尤其是对于人形机器人而言\cite{min2021development, melchiorri2013development}。较大的抓取力依赖于前臂肌肉提供动力支持，而精细操作则通过手掌的精确控制来实现。因此，仿生手的设计必须在力量输出与灵活性之间取得平衡，将前臂动力源与手掌精确性相结合，以实现高效、多功能的性能。这种分层驱动设计不仅更贴近人手的生理结构，而且为仿生手在实际应用中提供了一种可行的解决方案\cite{kim2014roboray, weng2025bidexhand}。

\begin{figure}
    \centering
    \includegraphics[width=1\linewidth]{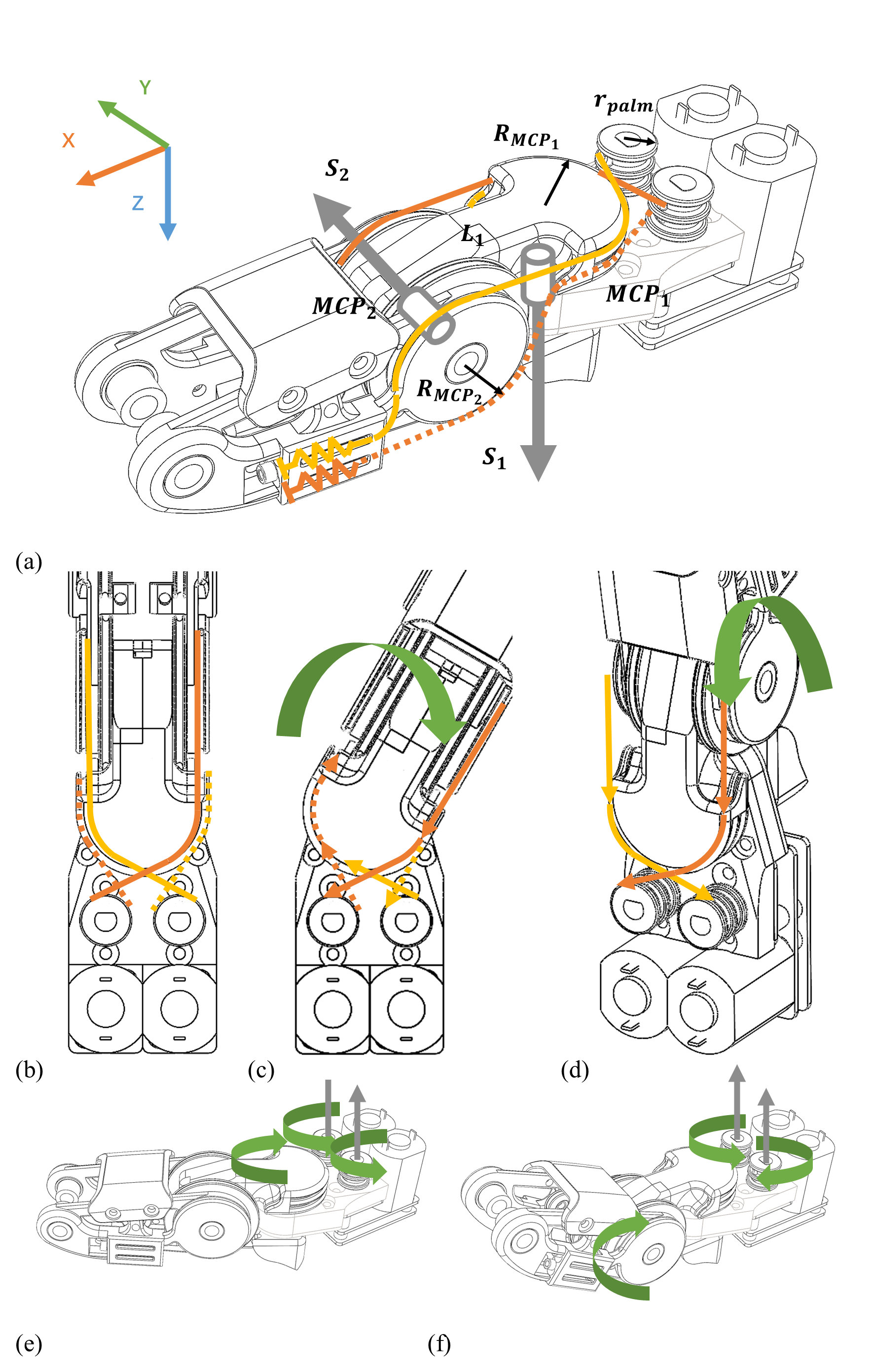}
    \caption{
    (a) Structure of MCP joint incorporating spring pre-tensioning mechanisms to compensate tendon slack (springs mounted at tendon termini). Dashed lines denote extension tendons, solid lines indicate flexion tendons; (b) Axial projection of MCP transmission structure, where each pulley module outputs two antagonistic tendons; (c)(e) Simultaneous counterclockwise rotation of dual motors causes synchronous contraction of the left antagonistic tendon pair, generating rightward deflection moment at MCP joint; (d)(f) Antagonistic motor rotation increases tension in solid tendons while decreasing dashed tendon tension, producing adduction moment at MCP joint.
    }
    \label{fig:mcp_joint}
\end{figure}

\section{Design}

% CAD design of the IRMV Hand. 图（a）展示了IRMV Hand的运动学结构，其具有15个运动自由度，与人类手掌运动自由度相似，末端DIP运动自由度可以根据需求设置为固定或运动。图（b）展示了IRMV Hand整体结构，其尺寸大约为100mm*100mm*265mm，指宽20mm，与成人手部尺寸相似。小臂部分部署了5枚用于驱动抓握功能的电机，掌部部署了10枚用于辅助操作任务的电机

% (a) MCP关节的传动结构，其中绳末端安装了弹簧做预张紧，虚线描述做伸指运动的绳腱，实线描述做屈指运动的绳腱; (b) MCP关节的传动结构的俯视图，其中每个转盘将传出两根方向相反的绳腱; (c)(e)两枚电机同时逆时针旋转时，指左侧两根绳腱被同时收缩，引起MCP 1向右弯曲；(d)(f)两枚电机异向旋转时，实线绳腱被收缩，虚线绳腱被松弛，引起MCP 2向内收缩。

\begin{figure*}
    \centering
    \includegraphics[width=1\linewidth]{hand.png}
    \caption{ CAD design of the IRMV Hand.  (a) illustrates the kinematic structure of the IRMV Hand, which possesses 15 DOF, similar to the DOF of a human hand. The DIP joint's DOF can be configured as either fixed or active based on requirements. (b) shows the overall structure of the IRMV Hand, with dimensions of approximately 100mm × 100mm × 265mm and a finger width of 20mm, closely resembling the size of an adult human hand. The forearm section houses five motors for driving grasping functions, while the palm section incorporates ten motors for assisting operational tasks.}
    \label{fig:hand}
\end{figure*}

\subsection{Mechanics}

% MCP关节设计

% (a) 结合弹簧预张紧机制以补偿肌腱松弛的MCP关节结构（弹簧安装在肌腱末端）。虚线表示伸肌腱，实线表示屈肌腱；(b) MCP传动结构的轴向投影，其中每个滑轮模块输出两根对立的肌腱；(c)(e) 双电机同时逆时针旋转导致左侧对立肌腱对同步收缩，在MCP关节产生向右的偏转力矩；(d)(f) 对立电机的旋转增加了实心肌腱的张力，同时降低了虚心肌腱的张力，在MCP关节产生内收力矩。

As shown in Fig. \ref{fig:mcp_joint}, the MCP joint's motion is designed with two separate and orthogonal rotational axes. This configuration facilitates the design of a precise cable-driven mechanism and supports a high-load-bearing structure. To address the transmission of the MCP joint, we developed a parallel cable-driven system, utilizing two motors to control its motion. The motors are mounted in the palm, mirroring the anatomical structure of a typical human hand. Each motor's output shaft is connected to a bi-directional pulley, controlling cables for both clockwise and counterclockwise motion. During finger abduction/adduction, bilateral coupled motors rotate synchronously: the left motor tensions solid-orange tendon while relaxing dashed-orange tendon, with the right motor executing identical actions to drive the MCP1 joint toward rightward deflection. Conversely, when the motors rotate in opposite directions, the MCP2 joint performs finger flexion or extension. Additionally, the transmission cables mounted on the forearm pass through the MCP2 joint, providing support to the joint.

% MCP的运动被设计成了分离的且互相垂直的两个旋转轴，这样的结构更有利于设计运动更为精准的绳传动方案，以及更高的承力结构设计。针对MCP关节的传动问题，我们设计了一种并联的绳驱动方式，采用两枚电机完成对MCP关节的运动控制。控制MCP关节运动的电机被安装在掌部，这与正常人类的生理结构类似。每枚电机的输出轴上将连接一个双向的转盘，提供顺时针与逆时针运动控制的绳。% 如图X所示，当执行指侧偏摆运动时，双侧耦合电机实施同向旋转：左侧电机收紧橙色实线肌腱(T1)同时释放橙色虚线肌腱(T2)，右侧电机同步执行相同模式，从而驱动掌指关节(MCP1)产生右向偏转。而电机呈现相异转动的时候，MCP的第二关节将会收缩或抬起。除此之外，安装在小臂上的传动线缆将会穿过MCP的第二关节，并为第二关节提供一定的支撑能力。

\begin{figure}
    \centering
    \includegraphics[width=1\linewidth]{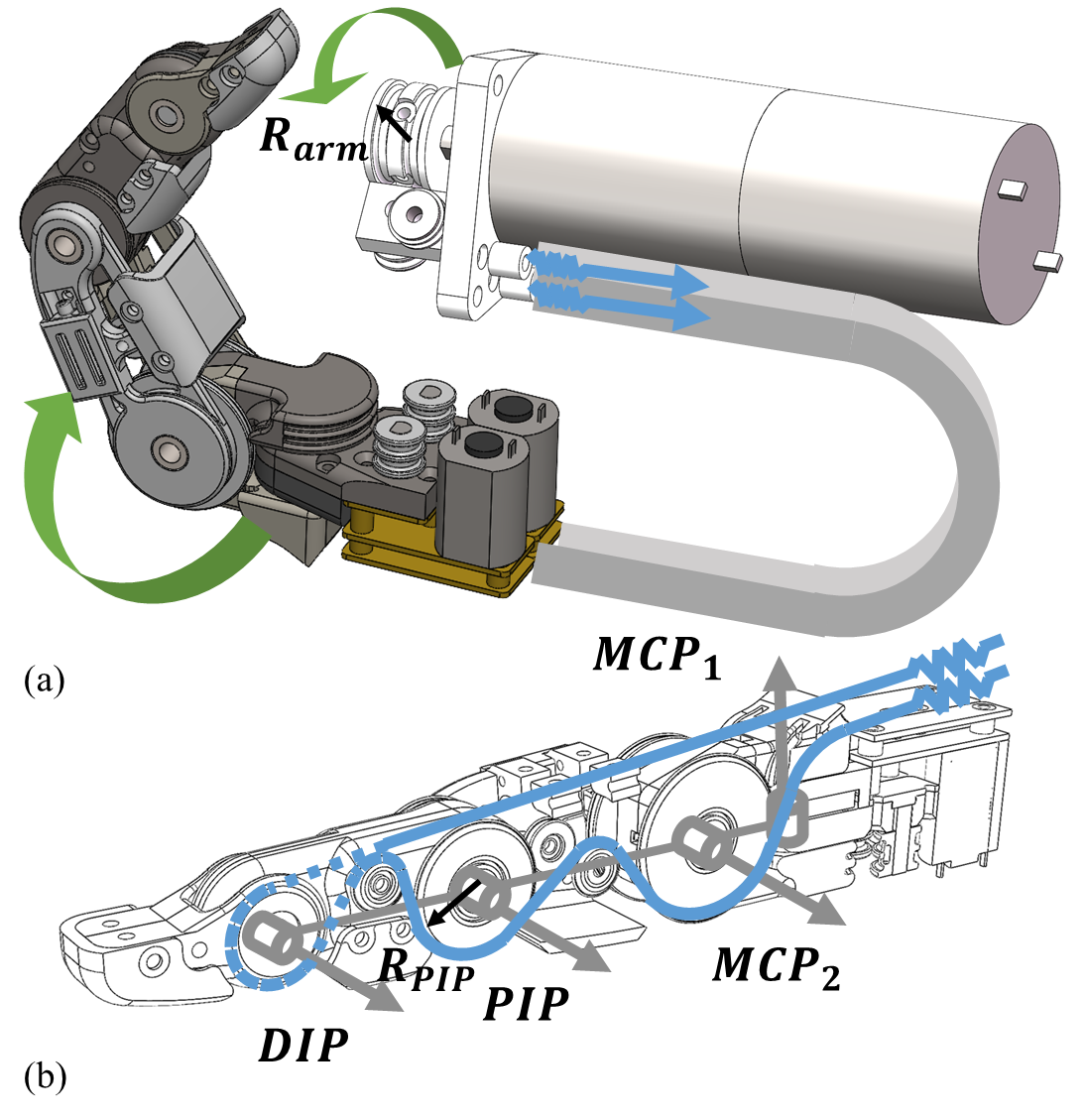}
    \caption{
    (a) Kinematic actuation of the PIP joint is driven by motor units located in the forearm compartment, achieving low-friction tendon force transmission via PTFE-lined conduits; (b) Tendon routing configuration for finger flexion and extension motions, generating directed moments at the MCP2 and PIP joints respectively
    }
    \label{fig:pip_joint}
\end{figure}

%PIP与DIP关节传动设计
In the human hand, the movements of the PIP and DIP joints are typically coupled. In this work, we have developed decoupled motion for the PIP and DIP joints. As shown in Fig. \ref{fig:pip_joint}, the cable driving the PIP joint passes through a pulley on the second MCP joint and terminates at the middle phalanx. Similarly, the cable driving the DIP joint follows a similar path, terminating at the distal phalanx. Both cables are routed through a conduit connected to the base of the finger, entering the forearm, and are controlled by motors located in the forearm. Since most bionic hand tasks only require actuation of the PIP joint, this design incorporates a dedicated actuation system for the PIP joint, while the DIP joint remains fixed. Critically, the preceding kinematic description represents specific operational cases. Flexion-extension and abduction-adduction motions across all joints achieve full decoupling control by computing motor target tensions from desired joint torque.

% 对于人手而言，通常PIP与DIP关节的运动是耦合的，这份工作中我们构建了运动解耦的PIP与DIP的运动。用于驱动PIP关节的绳缆会穿过安装在MCP第二关节的滑轮上，并终止在指中骨上。而驱动DIP关节的绳缆与之类似，最终中止在指尖骨上。控制两个关节的绳缆通过连接指根部的管道接入小臂，并由可以通过部署在小臂的电机进行控制。考虑到大部分仿生手任务仅需要PIP关节的传动，在本方案的整手设计中采用单独驱动PIP关节方案，而对于DIP关节进行固定。

This work utilizes DC motors as actuators, employing motors with varying power to meet the different drive requirements of the bionic hand. The MCP joint is driven by a small motor (N20) located in the palm. For grasping tasks, a higher-power motor (CHR-GM20-180) is employed, mounted in the forearm. In independently actuated bionic hands, transmission slack is a critical issue. To address this, a spring mechanism is positioned at the end of the cable to maintain tension during operation. The spring tensioning mechanism of the palm motor is located at the proximal phalanx, while that of the forearm motor is placed on the inner side of the forearm.

% 方案中采用直流电机作为驱动器，针对不同的驱动需求，仿生手采用不同功率的电机进行驱动。对于MCP关节采用小电机（N20）进行驱动，并安装在手掌中。对于抓握任务采用更高功率的电机（CHR-GM20-180）进行驱动，安装在小臂上。对于独立驱动式仿生手而言，传动松弛是一个极为致命的问题，因此在绳的末端部署上了弹簧结构，以保证传动过程中绳的紧绷状态。
The developed bionic hand exhibits similar dimensions and actuator distribution to those of an average adult male hand. For a complete single finger, the length of the first phalanx is 20mm, the second phalanx is 24mm, and the third phalanx is 32mm. The distance between the two MCP joints is 16mm, resulting in a total finger length of 92mm and a width of 18mm. The five fingers feature the same mechanical structure and are arranged in a manner similar to that of a human hand, with the thumb positioned horizontally. A total of 15 degrees of freedom have been designed, with 15 motors working in tandem to fulfill most manipulation and grasping tasks. The final constructed bionic hand has an overall length of approximately 265mm and a width of about 100mm.

% 本工作所开发的仿生手和普通成年男性的手具有相似的尺寸与动力源分布，对于完整的单根手指而言，第一指节长度20mm，第二指节长度24mm，第三指节长度32mm，MCP两个关节间距16mm，总计单指长度92mm，指宽18mm。五根手指采用相同的机械结构，并按类似于人手的位置进行排放，其中大拇指被横置。整体共计设计了15个运动自由度，并由15枚电机混合驱动，满足大部分操作与抓握的任务。最终构建的仿生手整体长度约265mm，宽约100mm

In this work, we adopt an independently actuated approach, combined with the concept of distributed driving, to design a cable-driven bionic hand.As showed in Fig. \ref{fig:hand}, the hand features five fingers, each with 3 degrees of freedom, aimed at mimicking the muscle distribution of the human hand while effectively reducing the overall structural size. The overall parameters of the IRMV Hand are described in Table \ref{tab:table1}.
% 在本研究中，我们采用了独立驱动式方案，结合分布式驱动的理念，设计出一种绳传动仿生手。该仿生手具备五根手指，每根手指具有3到4个自由度，旨在模仿人类手部的肌肉分布，并有效降低整体结构尺寸。

\begin{table}[H]
\caption{MECHANICAL AND ELECTRONIC SPECIFICATIONS OF THE IRMV HAND\label{tab:table1}}
\centering
\begin{tabular}{lc}
\toprule
\textbf{Quantity} & \textbf{Value} \\
\midrule
Number of fingers & five fingers \\
Finger Degree of freedom & 3 DoF \\
Hand Degree of freedom & 15 DoF \\
Dimension & 265 mm × 100 mm × 100 mm \\
Total weight & 1.4kg \\
Communication & USB \\
\bottomrule
\end{tabular}
\end{table}

\subsection{Electronics}

The control of a bionic hand system is a complex task that necessitates the stable operation of multiple motors and sensors. In most control tasks, force control methods enable more detailed feedback on the interaction with objects. The task will include torque control information for each finger joint, along with corresponding position feedback.
% 实现仿生手的电路控制任务是一个具有挑战性的工作，这需要实现大量电机与传感器的稳定运行。在大部分的仿生手控制任务中，通过力控制的方法可以反馈更多的操作物体信息。控制任务中将会描述手指每个关节的扭矩控制信息，并反馈得到对应的位置信息。

\begin{figure}[H]
    \centering
    \includegraphics[width=1\linewidth]{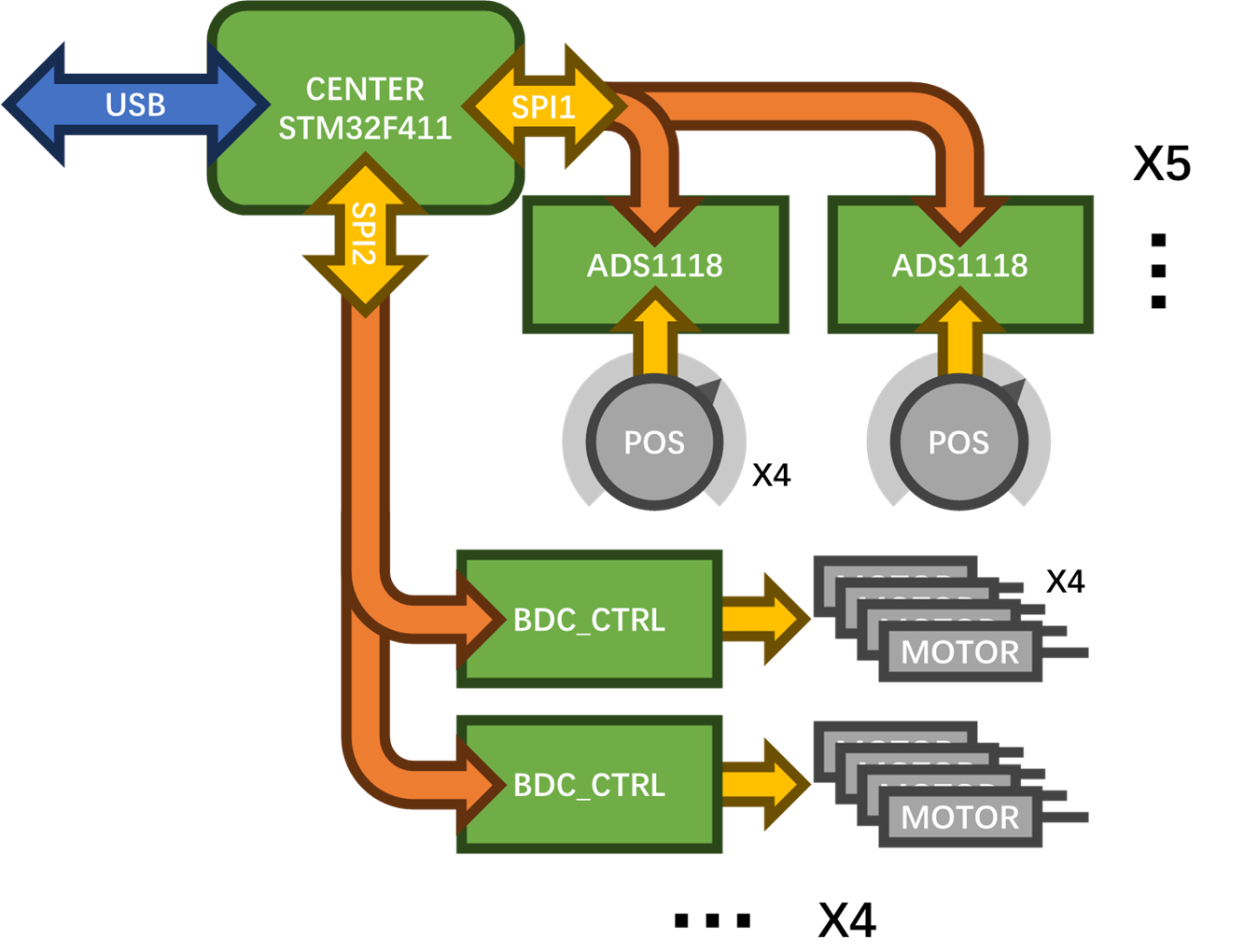}
    \caption{Hardware system of the IRMV Hand.}
    \label{fig:com_framework}
\end{figure}

The hardware system of the IRMV hand is shown in Fig. \ref{fig:com_framework}. The IRMV hand employs five position sensor boards and four motor driver boards, with the STM32F411 selected as the central controller. Communication with peripheral devices is established via two SPI channels, and the system is controlled by the host computer through a virtual USB serial port. The overall data communication frequency is set at 50Hz, allowing for the collection of position data and the control of motor currents to regulate the bionic hand’s output position. The central controller will reconfigure the sensor and drive interfaces for each finger, encapsulating them into torque control and position feedback for the respective finger joints.
% 仿生手共计使用五枚位置传感板和四枚电机驱动板，主控选择STM32F411作为控制核心，分别通过两路SPI控制从设备，并依靠虚拟USB串口受到上位机控制。整体数据通讯频率运行在50Hz，采集位置信息并控制电机电流值以控制仿生手输出位置。主控器中将重新配置每根手指的传感与驱动接口，并封装为对应指关节的扭矩控制与位置反馈。

\begin{figure}[H]
    \centering
    \includegraphics[width=1\linewidth]{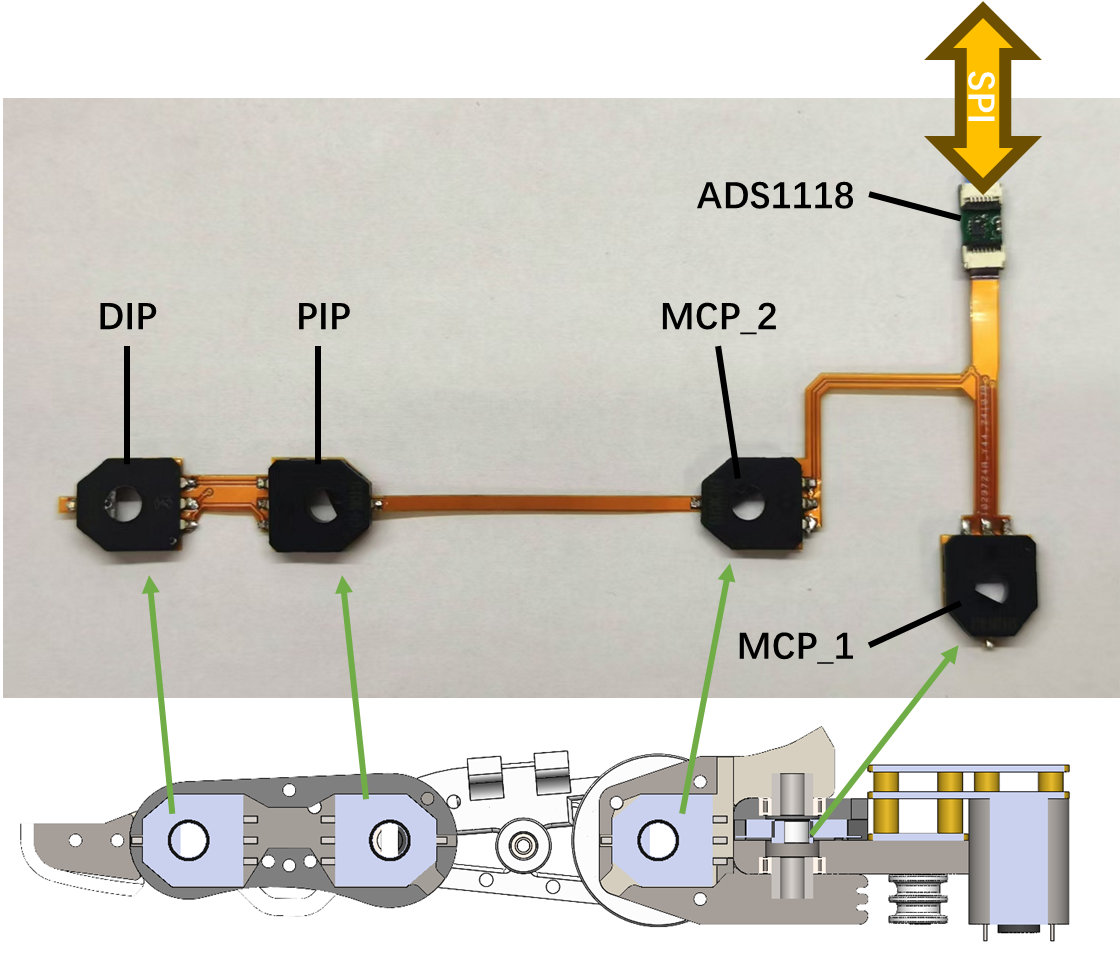}
    \caption{Position sensing circuit of the IRMV Hand.}
    \label{fig:ADS1118}
\end{figure}

To determine the joint angles of the bionic hand, each finger is equipped with four position sensors, totaling 20 sensors. Position sensing is achieved using commercial rotary potentiometers (SV01A103AEA01R00), connected via FPC flexible printed circuit boards and installed internally in each finger. An ADS1118 is placed at the base of each finger for data acquisition, with a maximum joint position sampling rate of 200Hz. \textcolor{red}{The ADS1118 analog-to-digital converter features a 16-bit ADC resolution, enabling the finger joints to achieve an angular resolution better than 0.02°.} This part is shown in Fig. \ref{fig:ADS1118}.
% 为了确定仿生手中所有的关节角度，每根手指安装四枚位置传感器，总计20枚。位置传感使用了商用旋转式电位计（SV01A103AEA01R00 ）实现，通过FPC软板进行连接，安装在每根手指的内部。每根手指尾部配置一枚ADS1118进行数据采集，最高可达到200HZ的位置关节采样率

\begin{figure}
    \centering
    \includegraphics[width=1\linewidth]{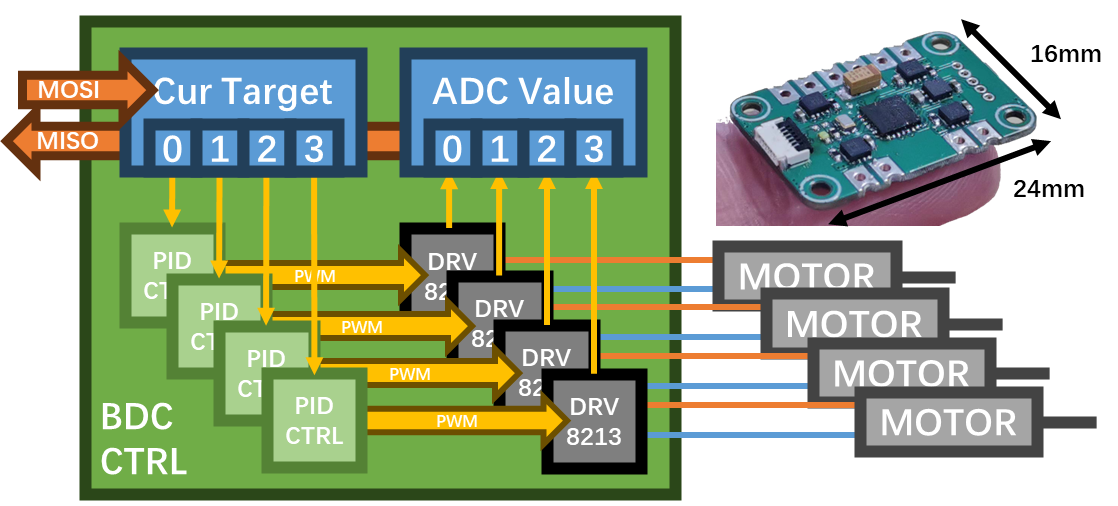}
    \caption{Brushed Motor Control Board of the IRMV Hand.}
    \label{fig:bdc_controller}
\end{figure}

Due to the large number of motors, we designed separate motor control boards(Fig. \ref{fig:bdc_controller}) to ensure stable and high-speed control of multiple motors. The motor driver board uses the STM32G071 as the main controller and the DRV8213 for motor driving. The control board operates at a frequency of 64 MHz and employs pulse width modulation (PWM) for motion control. The DRV8213 integrates a current sensing system and implements closed-loop PID current control. Communication is achieved via the SPI protocol, with CRC verification, supporting a maximum communication frequency of 1 kHz. The system can drive four motors with an adjustable voltage range of 3V to 12V. Additionally, to ensure the safety of movement, a low-pass filter is incorporated in the communication process to provide disconnection protection.
% 整体电机数量繁多，为了实现对多路电机的高速驱动控制任务，我们设计了单独的电机控制板以实现对多路电机的稳定高速控制。电机驱动板采用STM32G071作为主控制器，并使用DRV8213实现对电机的驱动。控制板运行频率在64Mhz，通过脉冲宽度调制技术实现对电机运动控制。DRV8213提供了集成的电流检测系统，并在系统中实现闭环PID电流环控制。控制板通过SPI协议受控，并依靠CRC进行校验，测试通信频率最高可达1khz。控制板可以实现3V-12V的任意电压的四路电机驱动。此外为确保运动的安全性，在通讯过程中设计了低通滤波实现断连的安全保护

\section{Analysis}

% 感觉没什么好写的，增加一节讨论指部运动控制好了

% This section provides a detailed analysis of the designed multi-degree-of-freedom bionic hand, focusing primarily on its mechanical structure and dynamic performance. The analysis is divided into several sub-sections, including kinematic analysis, mechanical analysis, and performance testing results with discussion.

\subsection{Kinematics and Dynamics}
\begin{figure}[H]
    \centering
    \includegraphics[width=1\linewidth]{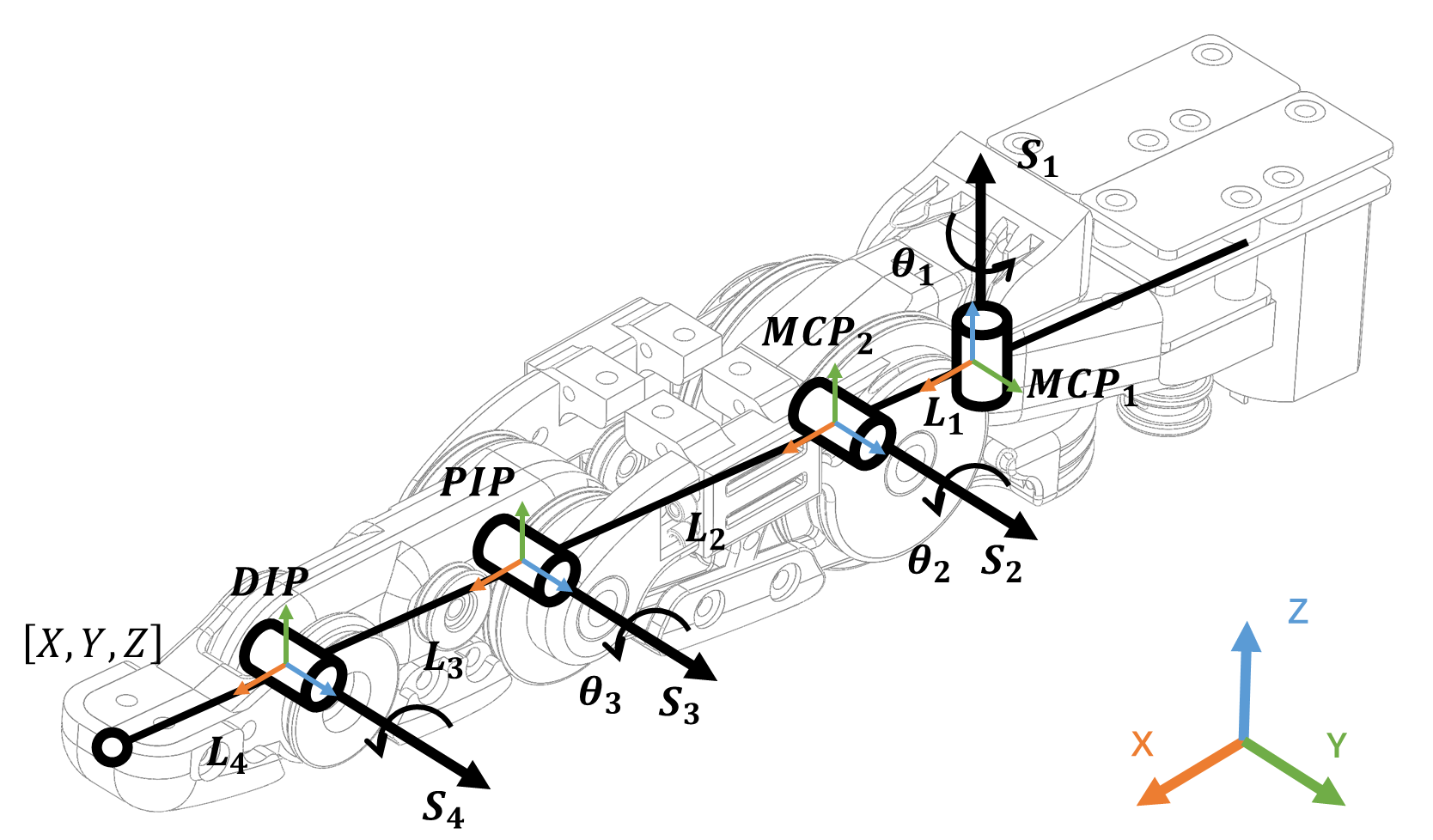}
    \caption{Kinematic Structure of a Single Finger}
    \label{fig:kinematic_01}
\end{figure}
% 如图所示，根据上述设计方案，我们定义了单根手指的正向运动学模型如下
As shown in Fig. \ref{fig:kinematic_01}, based on the structure above, we define the forward kinematic model of a single finger as follows
\begin{equation}
T(\theta) = \prod_{i=1}^3 e^{[S_i] \theta_i} M.
\label{eq:kine_01}
\end{equation}
% 锁定DIP运动关节后，通过正向运动学我们可以得到指尖坐标与指部关节角度的对应关系

After locking the DIP joint, the correspondence between the fingertip coordinates and the joint angles of the finger can be obtained through forward kinematics
\begin{equation}
\left\{
\begin{array}{ll}
&X s_1 - Y c_1 = 0 \\
&(L_2 + L_3 c_3) s_2 + L_3 s_3 c_2 - Z = 0 \\
&(\sqrt{X^2 + Y^2} - L_1)^2 + Z^2 - L_2^2 - L_3^2 - 2 L_2 L_3 c_3 = 0
\end{array}
\right.
\label{eq:kine_02}
\end{equation}
% 关节的位置由前文所述对应电机独立驱动，并分析关节扭矩与对应电机的扭矩输出。

% 如图X展示的绳腱传动结构可以分析出

%对于MCP指关节而言，其被两个相互垂直的旋转组成，并由掌间的电机进行驱动，而PIP关节则由小臂的电机直接驱动，所以手指关节对应的电机扭矩输出关系如下

The position of the joint is independently driven by the motors corresponding to the previously discussed configurations, and the relationship between joint torque and the torque output of the corresponding motors is analyzed. As show in Fig. \ref{fig:mcp_joint} and Fig. \ref{fig:pip_joint}, The MCP joint is driven by the motor located in the palm region, while the PIP joint is directly driven by the motor in the forearm. Therefore, the motor torque output relationship corresponding to the finger joints is as follows
\begin{equation}
\left\{
\begin{array}{ll}
    M_{MCP_1} = \frac{R_{MCP_1}}{r_{palm}} \left( \tau_{palm_1} + \tau_{palm_2} \right)\\
    M_{MCP_2} = \frac{R_{MCP_2}}{r_{palm}} \left( \tau_{palm_1} - \tau_{palm_2} \right)\\
    M_{PIP} = \frac{R_{PIP}}{r_{arm}} \tau_{arm}
\end{array}
\right.
\label{eq:kine_03}
\end{equation}

% 需要增加连接到前文

% 运动控制思路
% 整手的控制策略如图所示，首先依靠位置传感器采集到的位置信息计算位置差值，并通过动力学模型解算出各电机期望运动信息，依靠PID控制器计算对应电机目标电流。电机电流环控制集成在电机控制板中，依靠PID环进行调控电流输出，以控制扭矩，并设计有低通滤波器以保护设备断连而造成损坏

\subsection{Control Strategy}

\begin{figure}[H]
    \centering
    \includegraphics[width=1\linewidth]{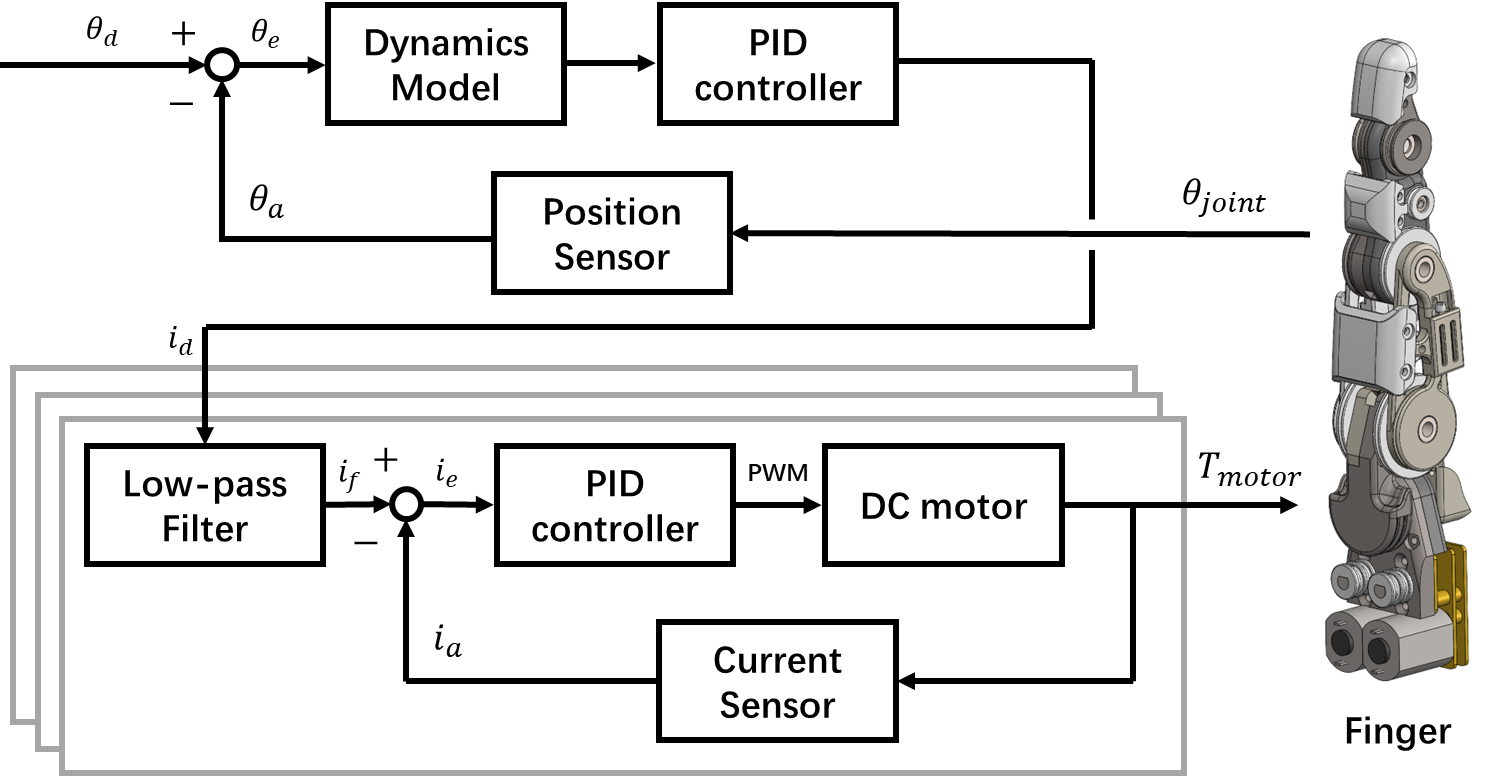}
    \caption{Control Block Diagram of a Single Finger}
    \label{fig:closed-loop}
\end{figure}

The control strategy for the entire hand is shown in the Fig. \ref{fig:closed-loop}. Initially, the position difference is calculated using the position data obtained from the position sensors, and the desired motion information for each motor is determined through a dynamic model. The target current for each motor is then calculated using a PID controller. The motor current control loop is integrated into the motor control board, where the PID loop regulates the current output to control the torque. A low-pass filter is also designed to protect the system from damage due to disconnection.

\section{Result}

\subsection{Performance of Single Finger}

\begin{figure}
    \centering
    \includegraphics[width=0.8\linewidth]{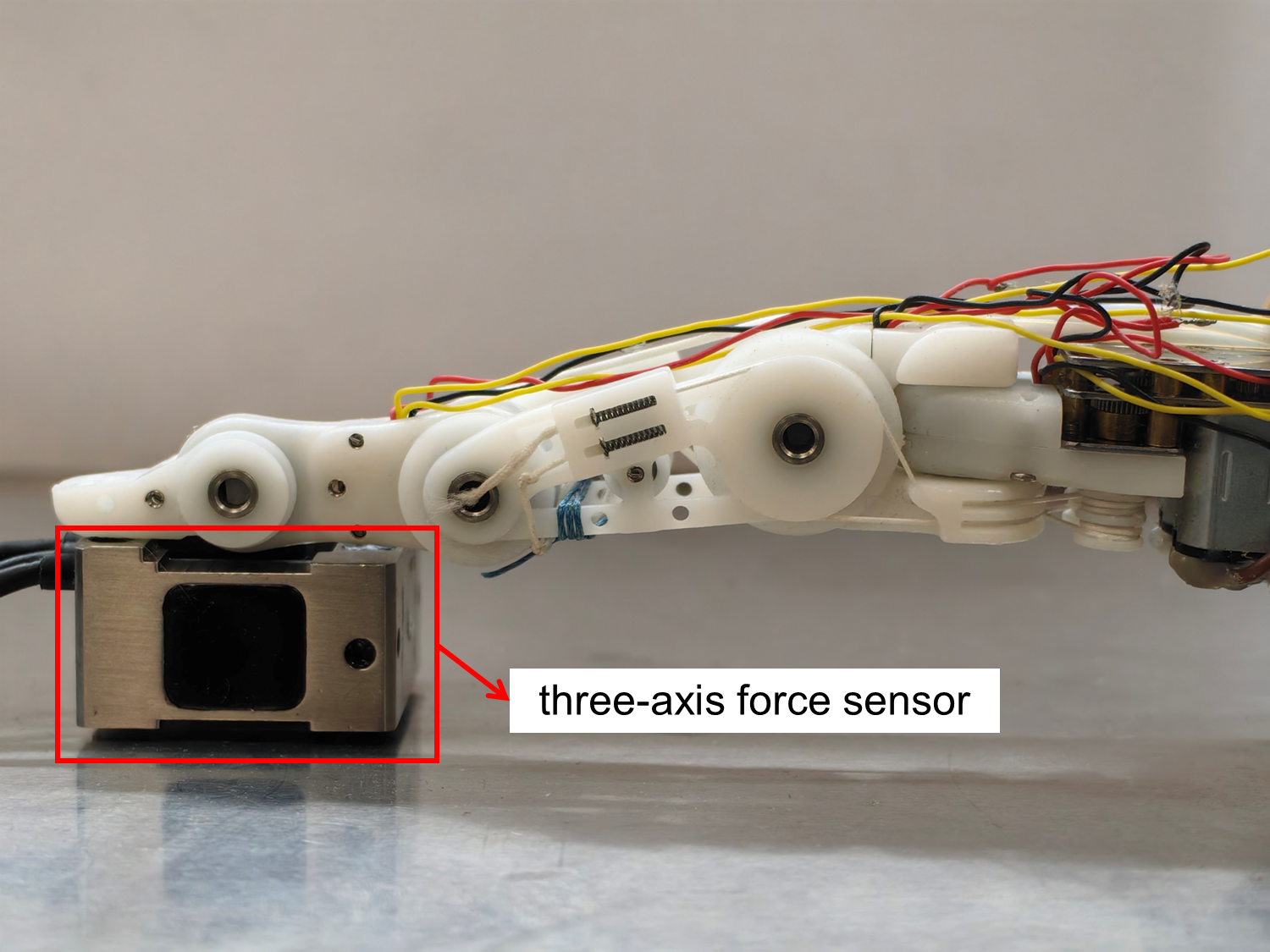}
    \caption{Testing Platform for contact force measurement.}
    \label{fig:result_real}
\end{figure}

\begin{figure}
    \centering
    \includegraphics[width=1\linewidth]{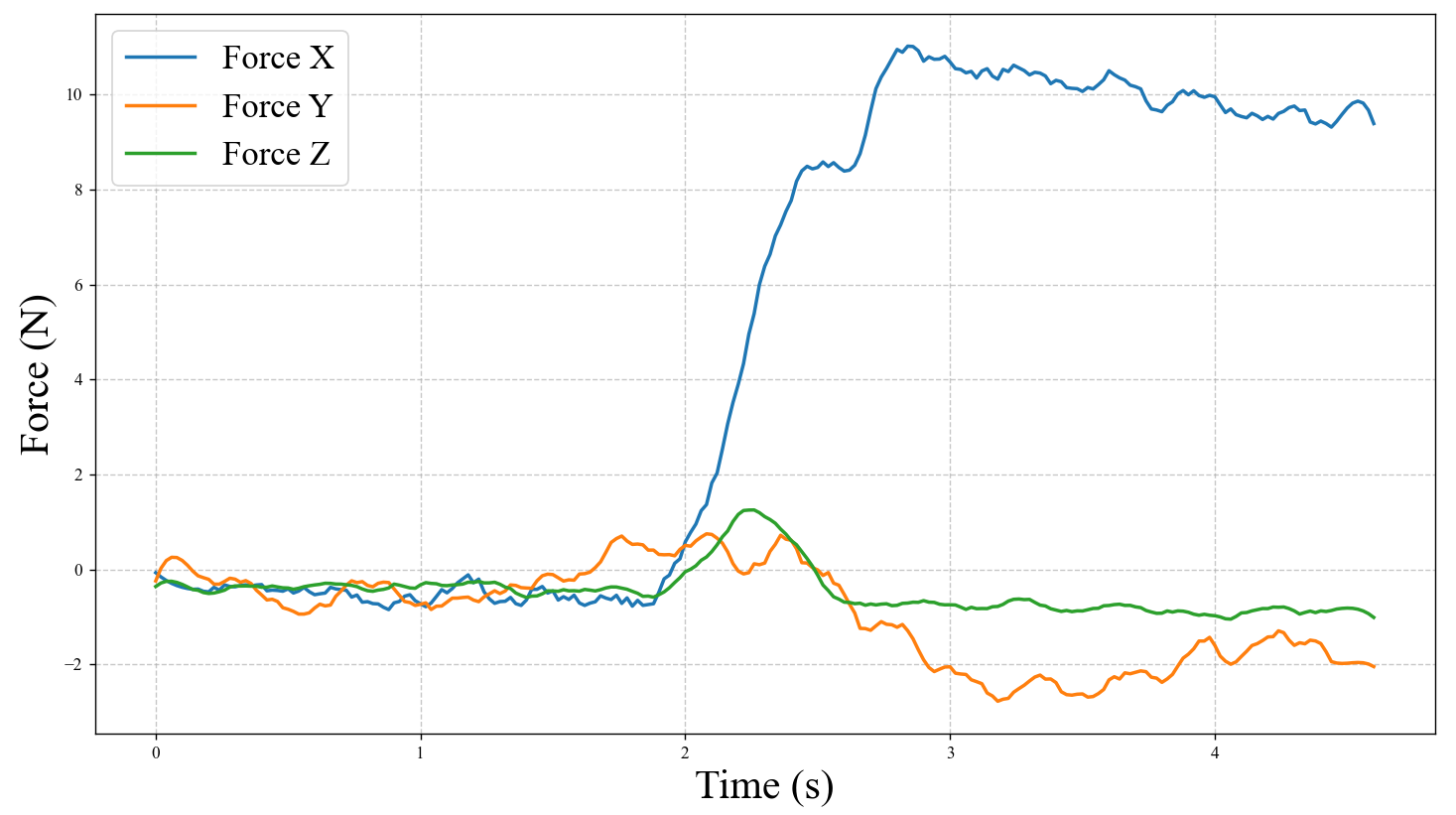}
    \caption{Contact force of single finger}
    \label{fig:result_exp}
\end{figure}

% 图X显示了Hand的每根手指与整体工作空间，所设计系统单指工作空间体积为
To evaluate the mechanical characteristics of fingertip compression, a dedicated testing platform was developed. As illustrated in     Fig. \ref{fig:result_real}, a three-axis force sensor (ZNSW-F) was rigidly mounted on the experimental platform to capture multidirectional force signals. Subjects were instructed to apply the maximum vertical compressive force using their index finger under controlled conditions. The force distribution across orthogonal axes is quantified in  Fig. \ref{fig:result_exp}. The maximum compressive force along the Z-axis reached 11 N, highlighting the dominant loading direction during fingertip compression. The design speed of the finger joint movement is approximately 30 rpm, and the design load for the whole hand grip is around 10 kg.
% 为了评估指尖压缩的力学特性，我们开发了一个专门的测试平台。如图\ref{fig:result_real}所示，一个三轴力传感器（ZNSW-F）被牢固地安装在实验平台上，以捕捉多向力信号。受试者被指示在受控条件下用食指施加最大的垂直压缩力。

% 图\ref{fig:result_exp}量化了正交轴上的力分布。沿Z轴的最大压缩力达到11牛顿，这突出了指尖压缩过程中的主要加载方向。
% 指关节运动速度设计在大约30rpm，整手抓握负载设计在10kg附近。

\subsection{Workspace}
\begin{table}[!t]
\caption{Finger Joint Motion and Length\label{tab:table2}}
\centering
\begin{tabular}{lcc}
\toprule
\textbf{Joint} & \textbf{Range of rotation} & \textbf{Phalange Length} \\
\midrule
MCP-1 & -30°-30° & 16mm \\
MCP-2 & 0°-90° & 32mm \\
PIP & 0°-90° & 24mm \\
DIP (Optional) & 0°-90° & 20mm \\
\bottomrule
\end{tabular}
\end{table}

\begin{figure}
    \centering
    \includegraphics[width=1\linewidth]{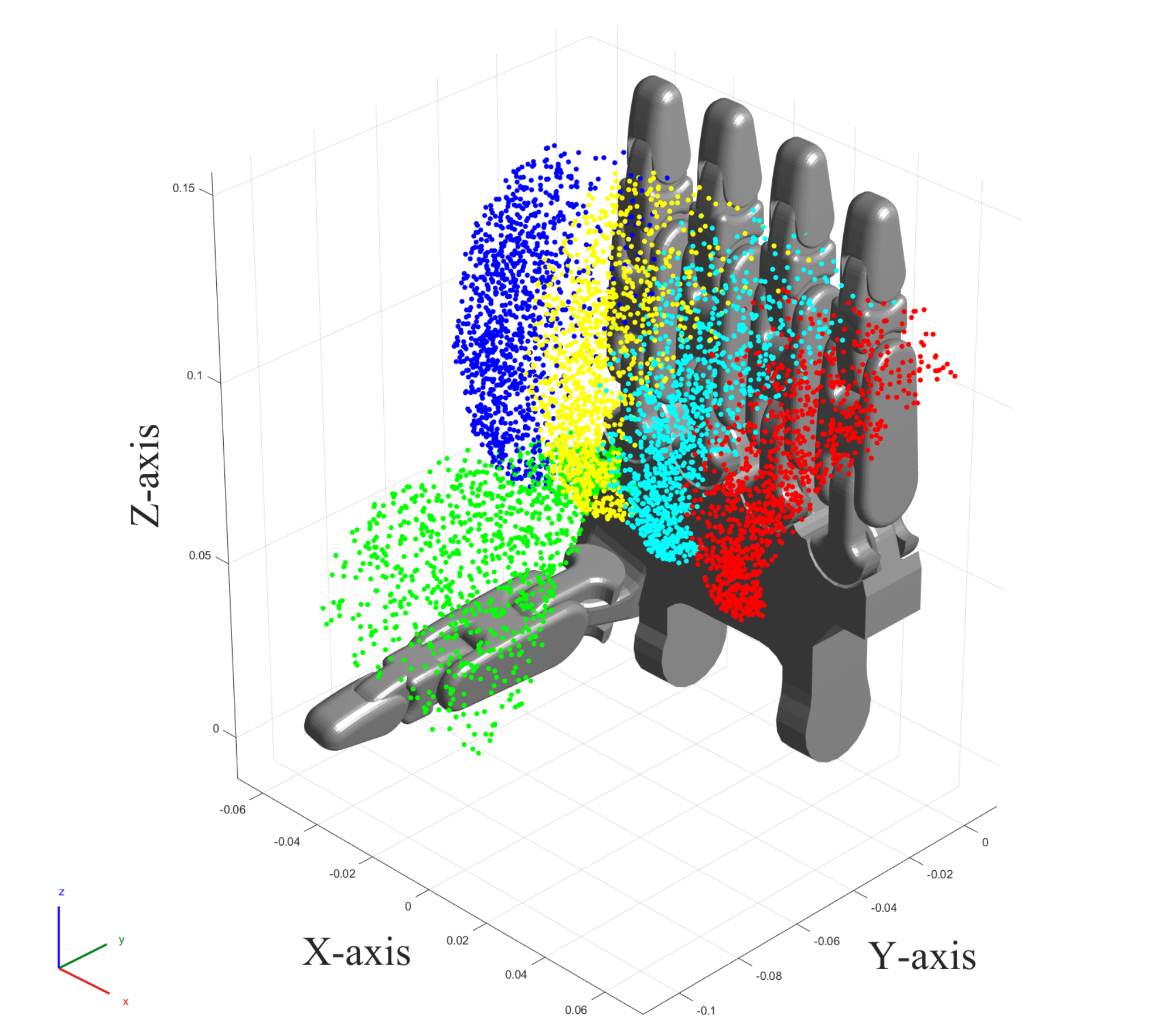}
    \caption{Workspace with Colored Points of the IRMV Hand.}
    \label{fig:workspace}
\end{figure}

Table \ref{tab:table2} shows the key parameters and range of motion of a single finger. Fig. \ref{fig:workspace} illustrates the total workspace of  each fingertip center point of the IRMV Hand. Each finger’s workspace is color-coded in the Fig. \ref{fig:workspace}, with different colors representing the motion space of each finger. The total volume of the workspace for each finger is 99 cm³, indicating the extent of motion flexibility provided by the system.
% 表\ref{tab:table2}展示了单个手指的关键参数和运动范围。图\ref{fig:workspace}展示了IRMV手部每个指尖中心点的总工作空间。在图\ref{fig:workspace}中，每个手指的工作空间都用不同颜色表示，不同的颜色代表每个手指的运动空间。每个手指的工作空间总体积为99立方厘米，这表明了该系统提供的运动灵活性程度。

\subsection{Object Grasping}
\begin{figure*}
    \centering
    \includegraphics[width=1\linewidth]{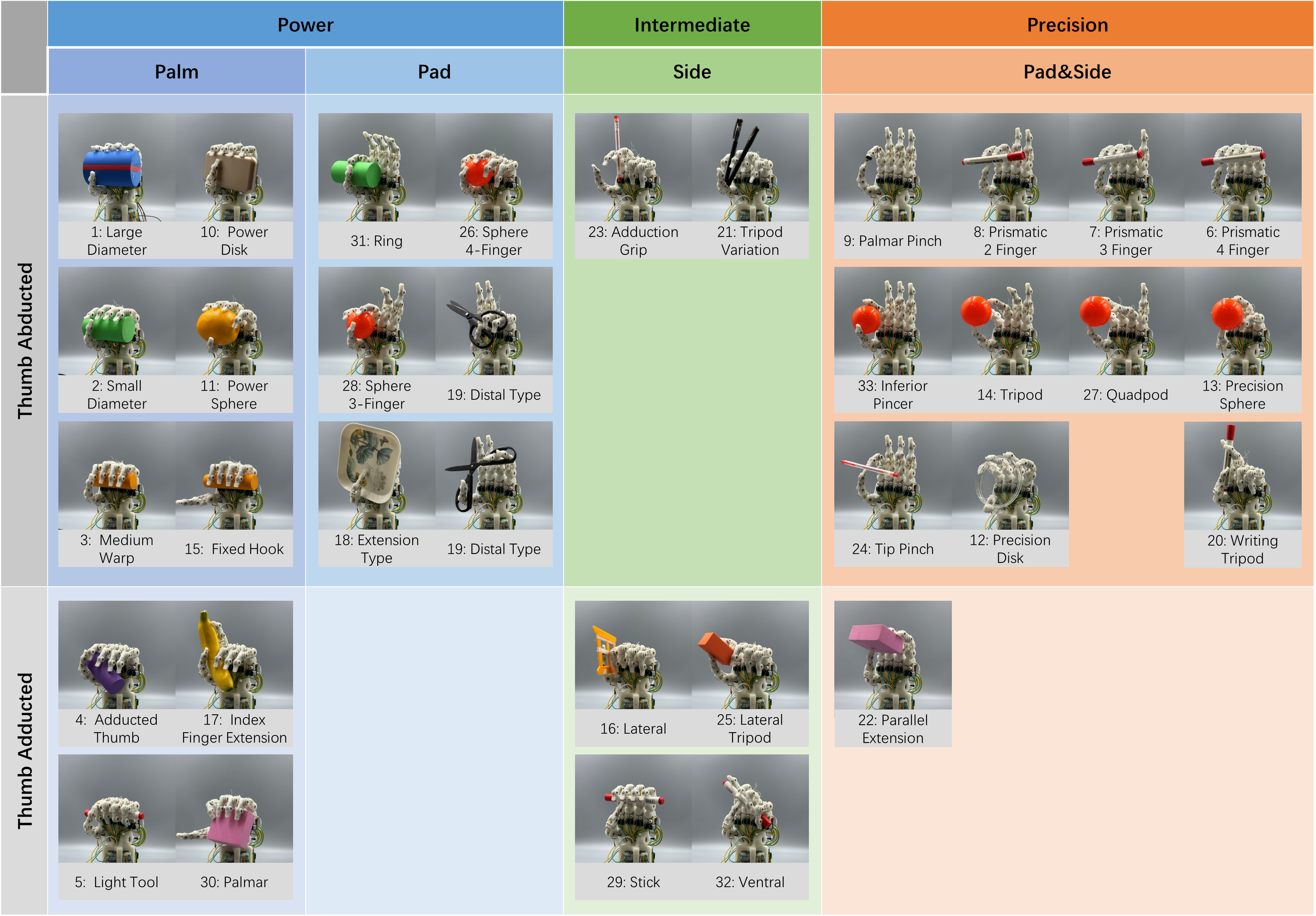}
    \caption{GRASP taxonomy on the IRMV Hand.}
    \label{fig:experiment}
\end{figure*}

    This study systematically validates the manipulation capability of the IRMV Hand using the Grasp Taxonomy benchmark established by Feix et al. \cite{feix2016grasp}. This benchmark evaluates the hand's adaptability to geometrically diverse objects by defining grasping paradigms under varying geometric and force constraints. As shown in Fig. \ref{fig:experiment}, the IRMV hand successfully achieved 33 distinct grasp types from the taxonomy, encompassing power grasps, precision grasps, and hybrid grasp modes. This empirically demonstrates its versatility and robustness in multi-scale object manipulation tasks.
% 我们采用Feix提出的Grasp分类基准对IRMV仿生手的操作能力进行验证，以呈现不同的抓取挑战，展示仿生手其适应不同形状和尺寸的能力。如图X所示，最终我们完成了分类基准中33中不同的握姿，验证了IRMV仿生手的广泛操作能力

% 我们在

The robotic hand was able to securely grasp these objects using different postures. The images highlight the hand’s capability to handle both symmetric and irregularly shaped objects, emphasizing its dexterity. The results confirm that the robotic hand can reliably grasp objects with a wide range of shapes, demonstrating its potential for numerous automation and robotic manipulation applications.
% 机械手能够以不同的姿势安全地抓取这些物体。这些图像凸显了机械手处理对称和不规则形状物体的能力，强调了其灵巧性。结果证实，机械手能够可靠地抓取各种形状的物体，展示了其在众多自动化和机器人操作应用中的潜力。

% 性能与评估指标 性能规格：虽然图11和图12有效地展示了夹持器的大工作空间和灵活性，但缺少定量的性能指标。作者应指明夹持器的指尖最大速度和能够成功操纵的最大物体重量。

\section{Conclusion}
This paper proposes a novel bionic hand structure that overcomes the limitations of traditional bionic hand designs, where the power sources are typically confined to the fingers, palm, or forearm. It addresses the issues of low power density and bulky forearm configurations. By adopting a distributed drive design, motors are strategically placed in the forearm and palm, optimizing space utilization while achieving a compact overall structure. The design offers 15 DoFs, making it closer to the physiological structure of the human hand. This design not only significantly improves the power density and motion performance of the bionic hand but also provides a new perspective for the development of high-performance bionic hands in the future.
% 本文提出了一种新型仿生手结构，克服了传统仿生手设计的局限性，即动力源通常局限于手指、手掌或前臂。该结构解决了功率密度低和前臂配置笨重的问题。通过采用分布式驱动设计，电机被巧妙地放置在前臂和手掌中，优化了空间利用率，同时实现了整体结构的紧凑化。该设计提供了15个自由度（DoF），使其更接近人手的生理结构。该设计不仅显著提高了仿生手的功率密度和运动性能，还为未来高性能仿生手的发展提供了新的视角。

Our next step involves expanding the sensory tasks of the bionic fingers, including the development of tendon tension sensors and tactile skin sensors. We also plan to improve the design of the wrist mechanism, with the goal of achieving more dexterous hand operation tasks.
% 我们的下一步工作是扩展仿生手指的感知任务，包括开发肌腱张力传感器和触觉皮肤传感器。我们还计划改进腕部机构的设计，目标是实现更灵巧的手部操作任务。

% 线缆张力（肌腱张力）管理：弹簧选型与张力控制：保持一致的线缆张力是一个关键挑战，尤其是在大旋转角度下。论文应详细阐述所选择的保持线缆张力的方法，包括弹簧的选型标准，以及该设计如何解决操作过程中潜在的张力变化问题。
% \begin{thebibliography}{99}

\bibliographystyle{IEEEtran}
\bibliography{hand_reference}
\nocite{*}

% \end{thebibliography}

\begin{IEEEbiography}[{\includegraphics[width=1in,height=1.25in,clip, keepaspectratio]{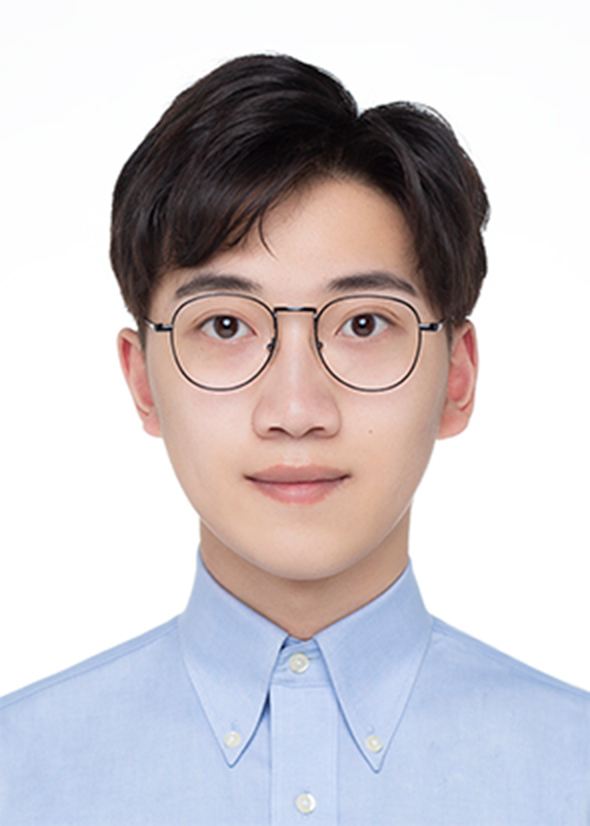}}]
    {Haoqi Han} received the B.Eng. degree in Mechatronics Engineering in 2018, from Beijing Institute of Technology, Beijing, China. He is currently working towards the Ph.D. degree in Computer Science and Engineering in Shanghai Jiao Tong University, Shanghai, China.

    His current research interests include biomimetic mechanisms. 
\end{IEEEbiography}

\begin{IEEEbiography}[{\includegraphics[width=1in,height=1.25in,clip, keepaspectratio]{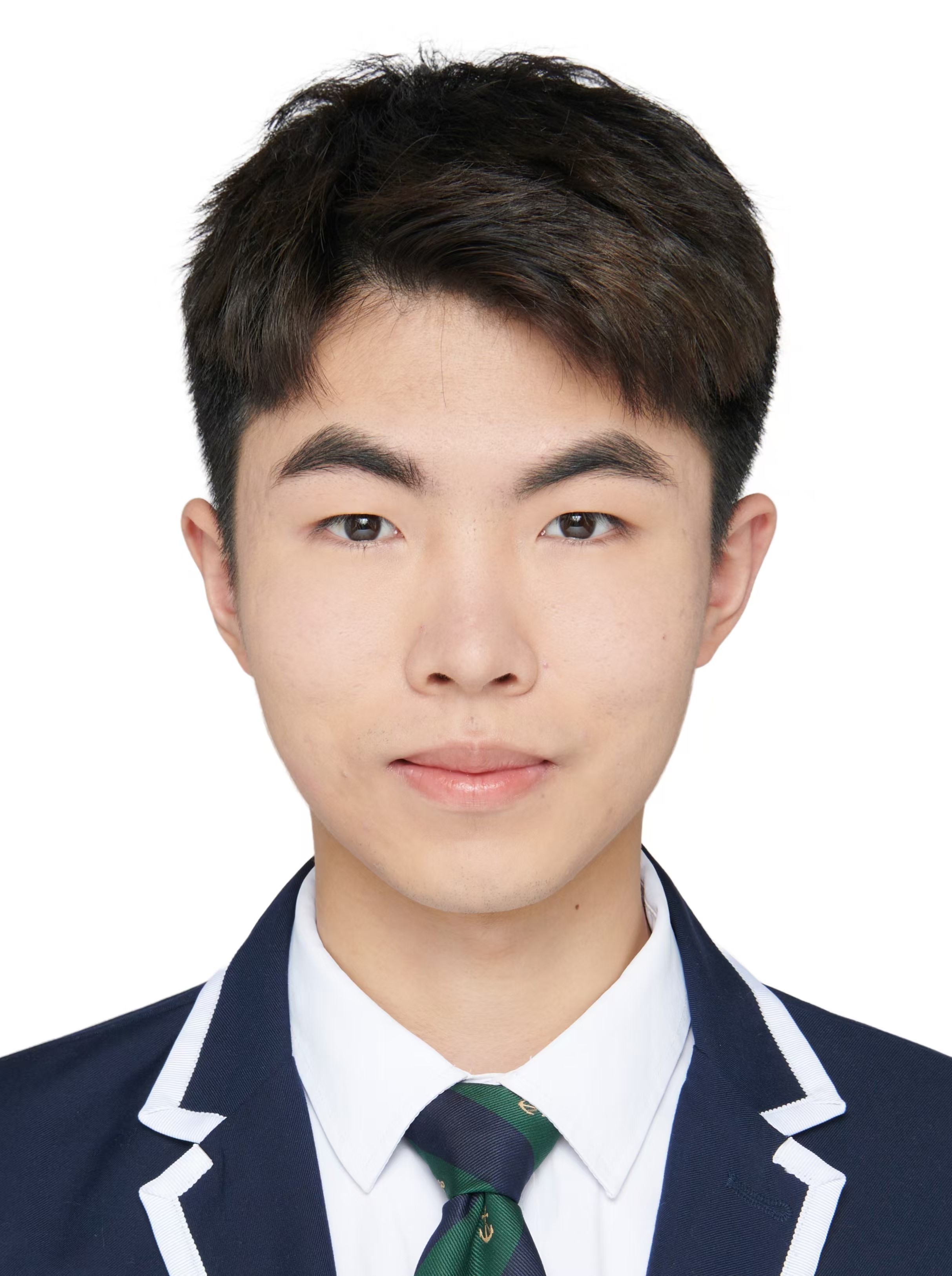}}]
    {Yi Yang} enrolled in the undergraduate program in Engineering Mechanics at Shanghai Jiao Tong University in 2023.
    
    His research interests include dexterous hand and robotics.
\end{IEEEbiography}

\begin{IEEEbiography}[{\includegraphics[width=1in,height=1.25in,clip, keepaspectratio]{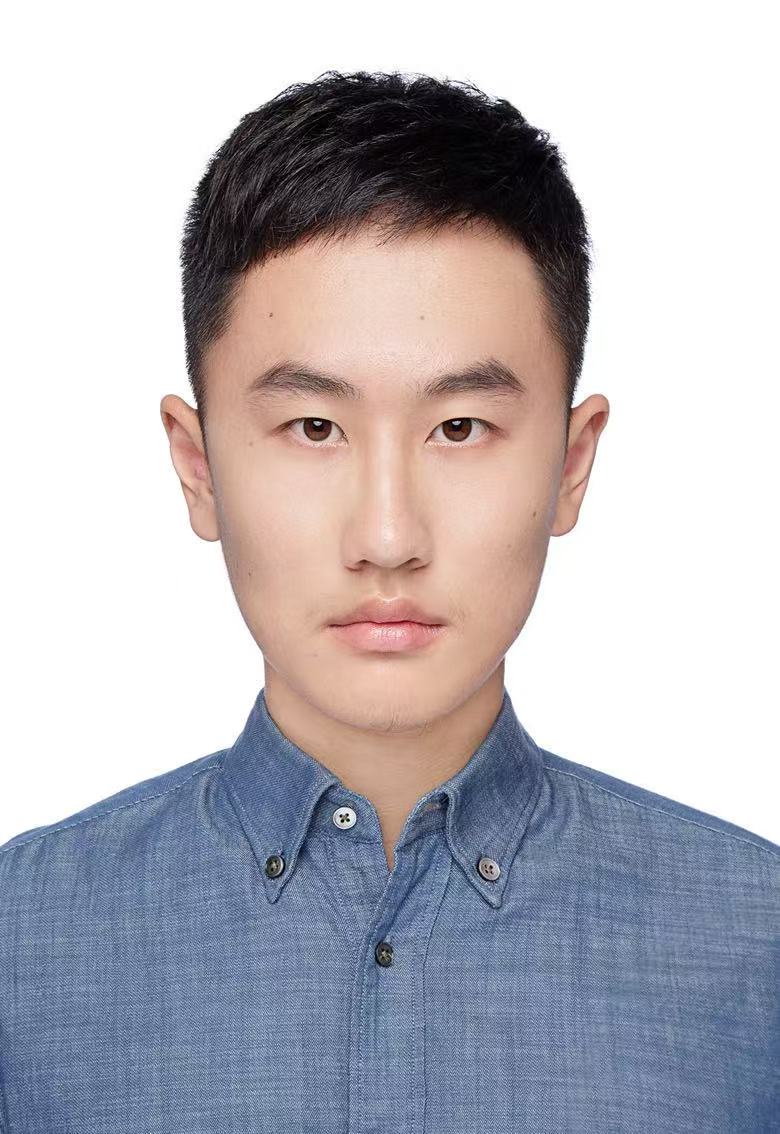}}]
    {Yifei Yu} enrolled in the undergraduate program in Automation Engineering at Shanghai university of electric power in 2023
    
    His research interests include mechanism design and robotics.
\end{IEEEbiography}

\begin{IEEEbiography}[{\includegraphics[width=1in,height=1.25in,clip, keepaspectratio]{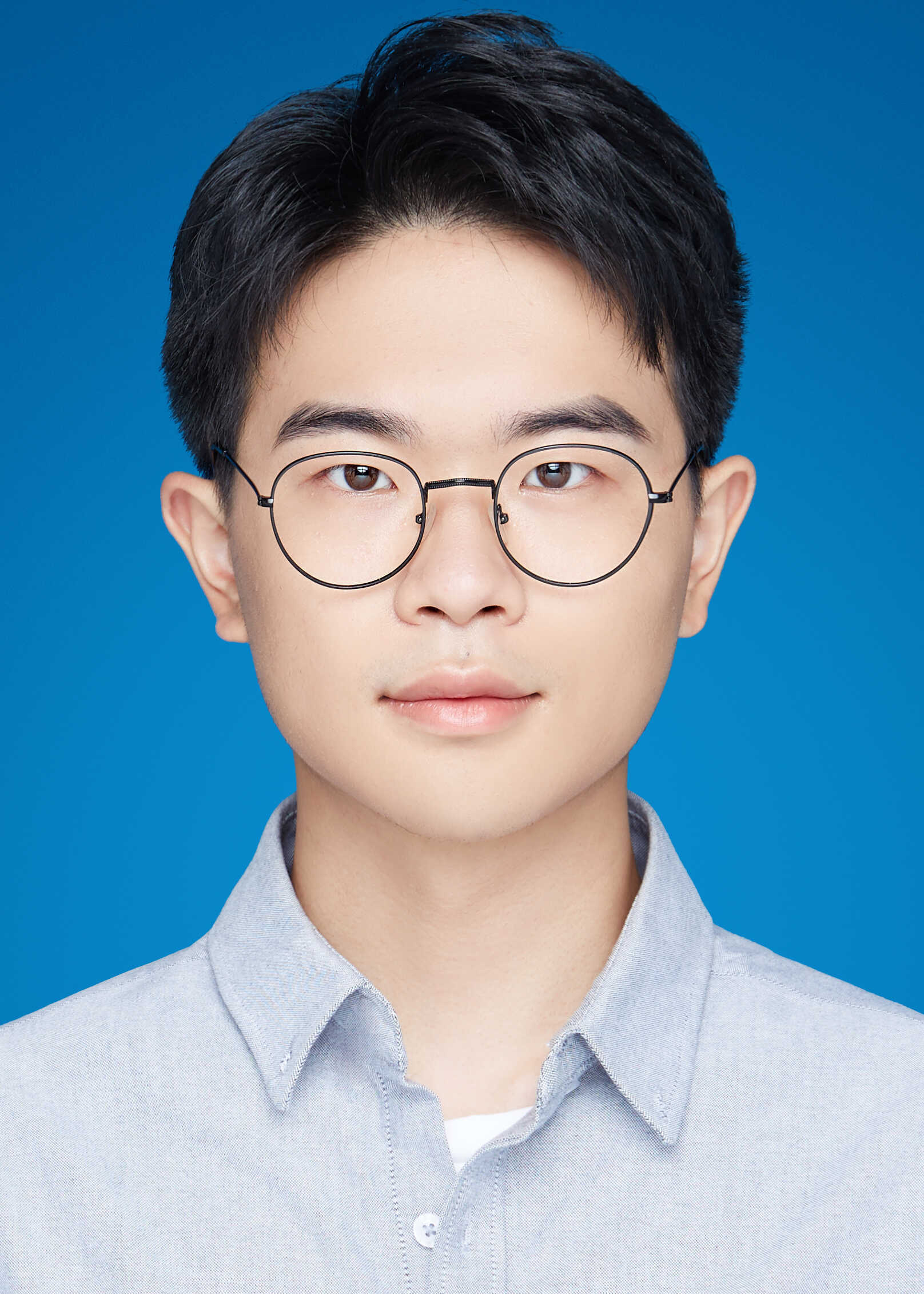}}]
    {Yixuan Zhou} received the B.Eng. degree in Mechatronics Engineering in 2018, from Beijing Institute of Technology, Beijing, China, where he also received an M.Eng. degree in Mechanical Engineering in 2021 too. He is currently working towards the Ph.D. degree in Control Science and Engineering in Shanghai Jiao Tong University, Shanghai, China.

    His current research interests include vision-based manipulation, visual servoing and trajectory planning.
\end{IEEEbiography}

\begin{IEEEbiography}[{\includegraphics[width=1in,height=1.25in,clip, keepaspectratio]{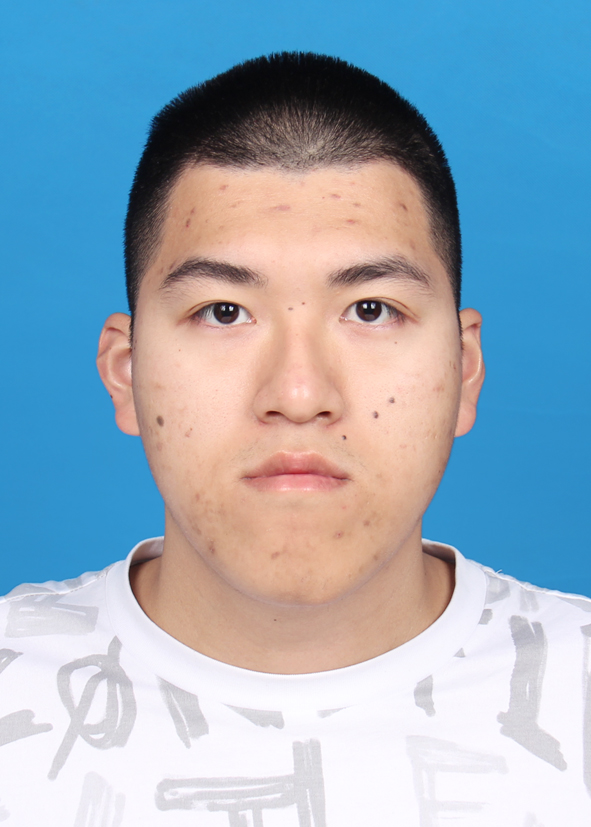}}]
    {Xiaohan Zhu} received the B.S. degree in automation from the Harbin Institute of Technology (HIT), Harbin, China, in 2023. He is currently working toward the M.S. degree in control technology and control engineering with the Department of Automation, Shanghai Jiao Tong University. 
    
    His research interests include soft robots, humanoid robots, and surgical robots.
\end{IEEEbiography}

\begin{IEEEbiography}[{\includegraphics[width=1in,height=1.25in,clip, keepaspectratio]{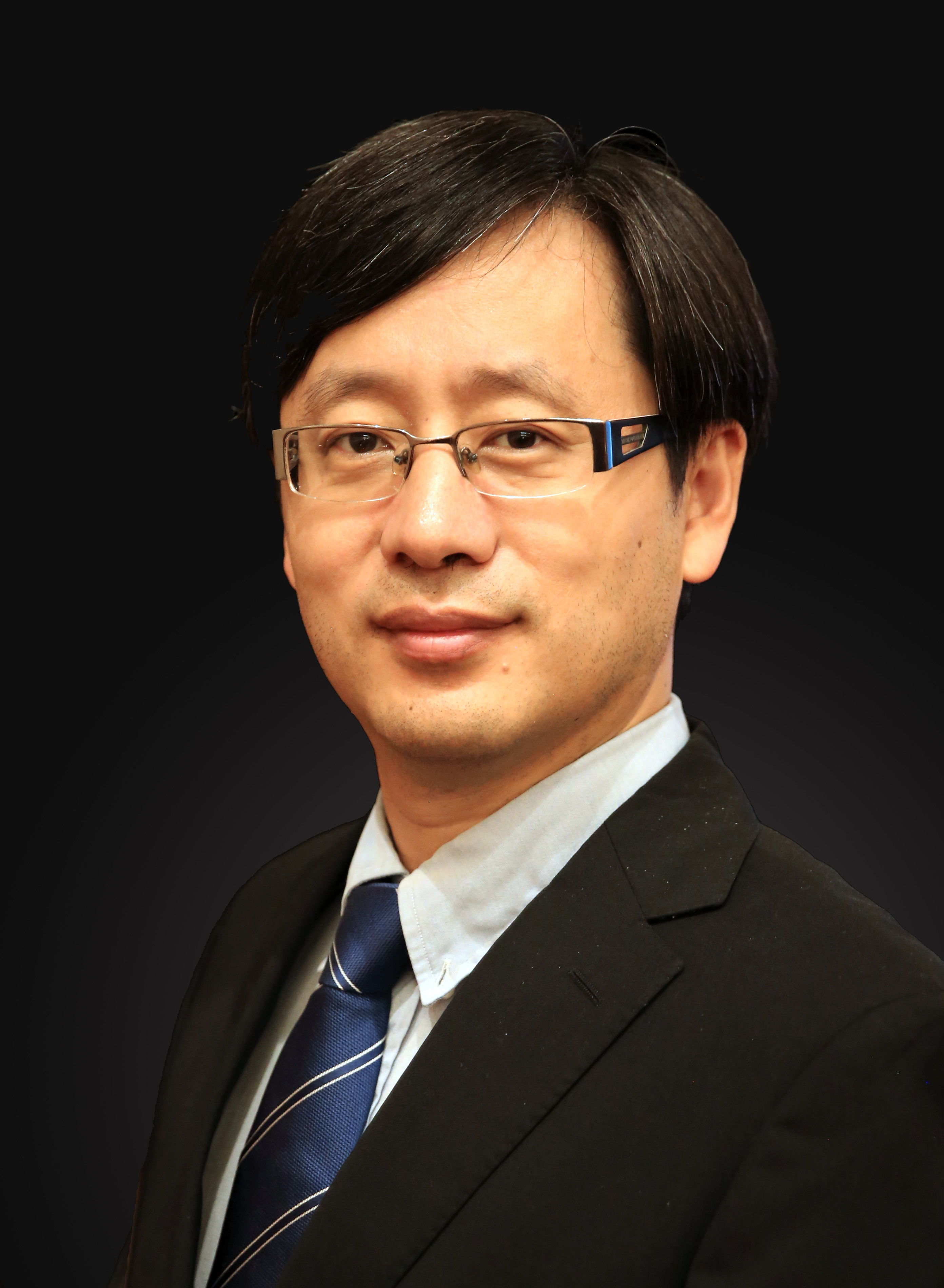}}]
    {Hesheng Wang} (SM’15) received the B.Eng. degree in electrical engineering from the Harbin Institute of Technology, Harbin, China, in 2002, and the M.Phil. and Ph.D. degrees in automation and computer-aided engineering from The Chinese University of Hong Kong, Hong Kong, in 2004 and 2007, respectively. He is currently a Distinguished Professor with the Department of Automation, Shanghai Jiao Tong University, Shanghai, China. His current research interests include visual servoing, intelligent robotics, computer vision, and autonomous driving. 

    Dr. Wang is an Associate Editor of Robotic Intelligence and Automation and the International Journal of Humanoid Robotics, a Senior Editor of the IEEE/ASME Transactions on Mechatronics, an Editor-in-chief of Robot Learning. He served as an Associate Editor of the IEEE Transactions on Robotics from 2015 to 2019, an IEEE Transactions on Automation Science and Engineering from 2021 to 2023, a Technical Editor of the IEEE/ASME Transactions on Mechatronics from 2020 to 2023, an Editor of Conference Editorial Board (CEB) of IEEE Robotics and Automation Society from 2022 to 2024. He was the General Chair of IEEE ROBIO 2022 and IEEE RCAR 2016, and the Program Chair of the IEEE ROBIO 2014 and IEEE/ASME AIM 2019. He will be the General Chair of IEEE/RSJ IROS 2025.
    %\\ \\ 
\end{IEEEbiography}

\end{document}